\documentclass[acmlarge, nonacm]{acmart}

\AtBeginDocument{%
  }

\usepackage{amsmath}

\usepackage{balance}       
\usepackage{graphics}      
\usepackage[T1]{fontenc}   
\usepackage{lipsum}
\usepackage{color}
\usepackage{booktabs}
\usepackage{textcomp}
\usepackage{microtype}        
\usepackage{ccicons}          
\usepackage{multirow}  
\usepackage{color,soul}  
\usepackage{subcaption}  
\usepackage{todonotes}
\usepackage{colortbl}  

\usepackage[export]{adjustbox} 

\usepackage{array}
\newcolumntype{P}[1]{>{\centering\arraybackslash}p{#1}}

\usepackage{arydshln}
\colorlet{tableheadcolor}{gray!25} 
\colorlet{tablerowcolor}{gray!15} 
\colorlet{tablerowcolor2}{gray!12} 
\colorlet{tablerowcolor3}{gray!25} 
\colorlet{tablerowcolor4}{gray!50} 
\newcommand{\rowcollight}{\rowcolor{tablerowcolor2}} %

\definecolor{ao(english)}{rgb}{0.0, 0.5, 0.0}

\newcommand{\edit}[1]{{\textcolor{black}{#1}}}
\newcommand{\darkgreen}[1]{\textbf{\sffamily{\textcolor{ao(english)}{#1}}}}

\usepackage{fancyhdr}   
\fancypagestyle{firststyle}{%
	\fancyhf{}
	\fancyfoot[C]{\textcolor{red}{This manuscript is under review. Please write to hharesamudram3@gatech.edu for up-to-date information.}}
}
\fancypagestyle{allstyle}{%
	\fancyfoot{}
	\fancyfoot[C]{\textcolor{red}{This manuscript is under review. Please write to hharesamudram3@gatech.edu for up-to-date information.}}
}

\setcopyright{acmcopyright}
\copyrightyear{2022}
\acmYear{2022}
\acmDOI{99.99/9999999}

\acmJournal{POMACS}
\acmVolume{99}
\acmNumber{99}
\acmArticle{99}
\acmMonth{99}

\begin{document}

\title{Towards Learning Discrete Representations via Self-Supervision for Wearables-Based Human Activity Recognition
}
\lfoot{\vfill \textcolor{red}{This manuscript is under review. Please contact harishkashyap@gatech.edu for up-to-date information.}}

\author{Harish Haresamudram}
\email{hharesamudram3@gatech.edu}
\affiliation{%
 	\institution{School of Electrical and Computer Engineering, Georgia Institute of Technology}
 	\city{Atlanta, GA}
 	\country{USA}
}

\author{Irfan Essa}
\email{irfan@gatech.edu}
\affiliation{
 	\institution{School of Interactive Computing, Georgia Institute of Technology}
 	\city{Atlanta, GA}
 	\country{USA}
}

\author{Thomas Pl\"{o}tz}
\email{thomas.ploetz@gatech.edu}
\affiliation{
 	\institution{School of Interactive Computing, Georgia Institute of Technology}
 	\city{Atlanta, GA}
 	\country{USA}
}

\renewcommand{\shortauthors}{Haresamudram et al.}

\begin{abstract}
	Human activity recognition (HAR) in wearable and ubiquitous computing is typically based on direct processing of sensor data streams.
	Sensor readings are translated into feature representations, either derived through dedicated preprocessing procedures, or integrated into end-to-end learning approaches.
	Independent of their origin, for the vast majority of contemporary HAR methods and applications, those feature representations are typically continuous in nature.
	That has not always been the case.
	In the early days of HAR, discretization approaches have been explored -- \edit{primarily motivated by the desire to minimize computational requirements on HAR, but also with a view on applications beyond mere activity classification, such as, for example, activity discovery, fingerprinting, or large-scale search.
	Those traditional discretization approaches}, however, suffer from substantial loss in precision and resolution in the resulting data representations with detrimental effects on downstream analysis tasks.
	Times have changed and in this paper we propose a return to discretized representations.
	We adopt and apply recent advancements in Vector Quantization (VQ) to wearables applications, which enables us to directly learn a mapping between short spans of sensor data and a codebook of vectors, where the index comprises the discrete representation, resulting in recognition performance that is generally on par with their contemporary, continuous counterparts -- sometimes surpassing them.
	Therefore, this work presents a proof-of-concept for demonstrating how effective discrete representations can be derived, enabling applications beyond mere activity classification but also opening up the field to advanced tools for the analysis of symbolic sequences, as they are known, for example, from domains such as natural language processing.
	Based on an extensive experimental evaluation on a suite of wearables-based benchmark HAR tasks, we demonstrate the potential of our learned discretization scheme and discuss how discretized sensor data analysis can lead to substantial changes in HAR. 	
\end{abstract}

%


\maketitle

\thispagestyle{firststyle} 
\pagestyle{allstyle}

\section{Introduction}
The widespread availability of commodity wearables such as smartphones and smartwatches, has resulted in increased interest towards their utilization for applications such as sports and fitness tracking (e.g., running, bicycling, and swimming) \cite{koskimaki2017myogym, ladha2013climbax, moller2012gymskill, khan2017activity}.
These devices benefit from onboard sensors, including Inertial Measurement Units (IMUs), which can track and measure human movements that are subsequently analyzed for understanding activities.
The ubiquitous nature of the devices, coupled with their form factor, enables the collection of large-scale movement data without substantial impact on user experience albeit without annotations. 

Human activity recognition (HAR) is one such application of wearable sensing, wherein features are extracted for segmented windows of sensor data, which are subsequently classified into specific activities of interest (or the null class).
The defacto approach for obtaining features is to compute them via statistical descriptors \cite{huynh2005analyzing} 
and the empirical cumulative distribution function \cite{hammerla2013preserving}, or instead to learn them directly from the data itself (e.g., via end-to-end training \cite{ordonez2016deep} or unsupervised learning \cite{plotz2011feature, saeed2019multi, haresamudram2020masked}). 
Either approach results in continuous-valued (or dense) features summarizing the movement present in the windows.

\edit{
	Alternatively, activity recognition has occasionally also been performed on discrete sensor data representations (e.g.,  \cite{stiefmeier2007gestures}).
	In those cases, short windows of sensor data are converted into \emph{discrete} symbols, where each symbol typically covers ranges of sensor values and a span of time.
	The motivation for such discretization efforts was to convert complex movements into a smaller, finite alphabet of discrete (symbolic) representations, thereby simplifying tasks such as  spotting gestures \cite{stiefmeier2007gestures}, and recognizing activities \cite{junejo2012using, sousa2018human} via the use of efficient algorithms from string matching and bioinformatics, or even simple Nearest Neighbors in conjunction with Dynamic Time Warping (DTW).
	Some approaches for deriving the small collection of symbols include Symbolic Aggregate approXimation (SAX) \cite{lin2007experiencing} and the Sliding Window and BottomUp (SWAB) algorithm \cite{keogh2001online, berlin2012detecting}.
	SAX in particular, is especially effective at discretizing even long duration time-series efficiently \cite{lin2007experiencing, stiefmeier2007gestures}.
}

\edit{Existing discretization methods are rather limited with regards to their expressive power -- resulting in substantial loss of resolution}, as movements and activities can only be expressed by a small \edit{alphabet of symbols}, \edit{which often} negatively impacts downstream recognition performance.
\edit{It is also observed} especially acutely in \edit{tasks} where minute differences in movements are important for discriminating between activities, e.g., in fine-grained gesture recognition. 
Moreover, the discretization methods are more difficult to apply to multi-channel sensor data, requiring specialized handling and containing exploding alphabet sizes \cite{montero2018mboss}.
\edit{
	Especially the low recognition accuracy coupled with difficulty in handling multi-sensor setups has resulted in discretization methods falling behind their continuous representation counterparts and, thus, have been somewhat abandoned. 
	 }
	 
	\edit{
	Yet, discrete representations--that are on par with contemporary continuous representations--can be crucial for tasks such as activity/ routine discovery via characteristic actions \cite{minnen2006discovering}, discovering multi-variate motifs from sensor data \cite{minnen2007discovering}, dimensionality reduction \cite{lin2007experiencing}, and performing interpretable time-series classification \cite{nguyen2018interpretable}, since these techniques require the simplification of time-series -- as they would be produced by discretization.
}
\edit{
	In recent years, the study and application of Vector Quantization (VQ) techniques to, for example, automatic speech recognition has resulted in the ability to learn mappings between the--continuous--audio and--discrete--codebooks of vectors, i.e., to map  short durations of raw audio to discrete symbols \cite{baevski2019vq,baevski2020wav2vec}.
	In this paper, we propose to adopt and adapt such recent advancements of discrete representation learning for the HAR community, so that the symbolic representations of movement data can be derived in an unsupervised, data-driven manner, and be used for effective sensor-based human activity recognition tasks. 
}

\edit{
	To this end, we apply \textit{learned} Vector quantization (VQ)\cite{van2017neural, baevski2019vq} to wearables applications, which enables us to directly \emph{learn} the mapping between short spans of sensor data and a codebook of vectors, where the index of the closest codebook vector comprises the discrete representation (also referred to as the codeword). 
	In addition, we utilize self-supervised learning as a base (via the Enhanced CPC framework \cite{haresamudram2023investigating}), thereby deriving the representations without the need for annotations.
}
\edit{
Sensor data are first encoded using convolutional blocks, which can handle multiple data channels (e.g., x-y-z axes of triaxial accelerometry) in a straightforward manner. 
This is followed by the vector quantization module which replaces the encoding with the nearest codebook vector. 
They are subsequently summarized using a causal convolutional encoder, which utilizes vectors from the previous timesteps in order to predict multiple future timesteps of data in a contrastive learning setup.
}

\edit{
	We present our method as a \emph{proof-of-concept} for discrete representation learning, and, as such, as a proposal for a return to discretized representations in HAR.\
	We focus on recognizing human activities in order to demonstrate the efficacy of the representations using a standard classification backend, yet we also outline the potential the proposed return to discretized processing has for the field of sensor-based HAR.
	The representations are derived using wrist-based accelerometer data from Capture-24 \cite{chan2021capture, gershuny2020testing, willetts2018statistical} -- a large-scale dataset that was collected in-the-wild and, as such, is representative for real-world Ubicomp applications of HAR. 
}
\edit{
	The performance of our discrete representation learning is contrasted against other representations, including end-to-end training (where the features are learned for accurate prediction) and self-supervised learning (where unlabeled data are first used for representation learning), the latter recently having seen a substantial boost in the field.
	The evaluation is performed on six diverse benchmarks, containing a variety of activities including locomotion, daily living, and gym exercises, and comprising different numbers of participants and three sensor locations (as detailed in \cite{haresamudram2022assessing}). 
}

\edit{
The conversion of continuous-valued sensor data to discrete representations often results in comparable activity recognition accuracy, which we show in our extensive experimental evaluation. 
In fact, in some cases the change in representation actually leads to improved recognition accuracy.
In addition to standard activity recognition, the return to--now much improved--discretization of sensor data also bears great potential for a range of additional applications such as activity discovery, activity summarization and fingerprinting, which could be used for large scale behavior assessments both longitudinally or population-wide (or both).
Effective discretization also opens up the field to the potential of entirely different categories of subsequent processing techniques, for example NLP-based pre-training such as RoBERTa \cite{liu2019roberta}--an optimized version of BERT \cite{devlin2018bert}--so as to further learn effective embeddings, improving the recognition accuracy.
}

The contributions of our work can be summarized as follows:
\begin{itemize}
	\item \edit{We combine learned Vector Quantization (VQ)--based on state-of-the-art self-supervised learning methods--with wearables-based human activity recognition in order to learn discrete representations of human movements.}
	
	\item We establish the utility of \edit{learned} discrete representations towards recognizing activities, \edit{where they perform comparably or better than state-of-the-art learned representations} on three datasets, across sensor locations.
	
	\item We also demonstrate the applicability of highly effective NLP-based pre-training (based on BERT \cite{devlin2018bert, liu2019roberta} \edit{and RoBERTa \cite{liu2019roberta}}) on the discrete representations, which results in further performance improvements for all target scenarios. 
	
	\item \edit{We discuss the potential impact the new learned discrete representations have beyond standard activity recognition, outlining application cases that are enabled by our new approach.}
\end{itemize}

\section{Background}
\label{sec:related}
As the goal of this work is to demonstrate the effectiveness of \emph{learning} discrete representations of sensor data, we first discuss previous methods for deriving the symbols, followed by techniques from other domains that learn discrete representations.
Finally, we discuss self-supervised methods for wearable sensor data, which can be used in conjunction with vector quantization for learning discrete representations.  

\subsection{Discretizing Sensor Data and Deriving Primitives}
Discretization of time-series data has traditionally been performed using computational methods such as Symbolic Aggregate approXimation (SAX) \cite{lin2007experiencing, shieh2008sax}.
In this technique, Piecewise Aggregate Approximation (PAA) is first utilized to obtain representations covering spans of time, which are subsequently symbolized using the alphabet size (which is a parameter to be tuned).
This process is simple, yet highly effective and fast for even long duration time-series data.
While originally proposed for single time-series, it has been extended to multi-channel data as well, by applying SAX separately to each channel and combining the tuples or first applying Principal Component Analysis (PCA) to reduce to a single channel \cite{mohammad2014robust}.
Other variations include applying Tf-idf weighting of the SAX features, as explored in SAX-VSM \cite{senin2013sax}.
Such computational discretization methods have also been applied for HAR applications, using SAX \cite{junejo2012using} and its variants \cite{sousa2018human}.

Multivariate Bag-Of-SFA-Symbols (MBOSS) \cite{montero2018mboss} also learns symbolic representations but via Symbolic Fourier Approximation (SFA) \cite{schafer2015boss, schafer2016scalable}. 
It adapts Bag-Of-SFA-Symbols in Vector Space (BOSS VS), which functions on univariate time-series to triaxial accelerometry. 
SFA utilizes the Discrete Fourier Transform (DFT) to obtain the coefficients for the data, after which Multiple Coefficient Binning
(MCB) is applied separately to obtain the symbols for each coefficient. 
The SFA word then consists of the tuple of symbols across the coefficients, and a histogram of these words across a window comprises the representation.
MBOSS simply stacks together the BOSS histograms for each channel separately, thereby resulting in the representation for each window, which is classified using simple classifiers such as kNN and Support Vector Machines (SVM).

An extension for the SFA method was proposed in Word ExtrAction for time SEries cLassification (WEASEL) \cite{schafer2017fast}.
It applies the ANOVA f-test to determine the most informative Fourier coefficients and subsequently applies information gain binning for determining the boundaries. 
Subsequently, feature selection is performed to only pick the relevant features without negatively impacting performance.
Many of these methods are surveyed in \cite{sousa2018human} in order to compare the raw performance as well as other factors such as the time taken for computation etc.

The Sliding Window and BottomUp (SWAB) algorithm \cite{keogh2001online, berlin2012detecting} has also been used to discretize sensor data, with the goal of detecting leisure activities using dense motif discovery.
In particular, \cite{berlin2012detecting} proposes an inference system for mood disorder research, where accurately recognizing specific leisure activities is of vital importance. 
Piecewise Linear Approximation first produces linear segments from sensor data, following which the slope between consecutive segments is binned to obtain the discrete representations. 
Suffix trees are utilized for extracting motifs that represent activities.

Activity discovery was explored in \cite{minnen2006discovering}, which involves the identification of activities from sensor streams. 
The aim was to discover recurring patterns, i.e., motifs, without annotations or segmentation, as they are statistically unlikely to occur and thus correspond to exemplar actions for activities.
Discovering motifs for time-series is challenging as they can be sparsely distributed, vary in duration, and exhibit some level of time-warping.
SAX is first employed to locally discretize the data, aiding in the discovery of motifs.
It was also employed in \cite{minnen2007improving} for improving the discovery of activities, as well as \cite{minnen2007detecting} for discovering motifs in multi-variate data.

Online gesture spotting has also been performed with discretized movements using string matching algorithms \cite{stiefmeier2007gestures}.
First, motion is represented by a string of symbols and efficient string matching methods  (incl. approximate matches) are employed to recognize gestures. 
Similarly, primitives of motion have been derived via shapelets \cite{ye2009time} for time series classification \cite{mueen2011logical}, wherein each shapelet is a local pattern highly indicative of a class.
They have been applied to trajectory classification as well, with the introduction of `movelets' \cite{ferrero2018movelets}.
Another technique for discovering primitives includes utilizing the matrix profile (and its extensions) \cite{yeh2016matrix, zhu2017matrix}, which also facilitates motif discovery.
As mentioned previously, the computational methods detailed above have lower performance relative to deep learning in general, and do not handle multi-channel data well.
Due to these issues, they have not been studied extensively in recent years.

\subsection{Discrete Representations Learning in Other Domains}
Learning discrete representations with deep networks was first introduced in an autoencoder setting (so-called Vector Quantized Variational AutoEncoder (VQ-VAE)) \cite{van2017neural}, where the encoder outputs discrete codes instead of continuous latents. 
This was achieved with the use of an online K-means loss, which allowed for a differentiable mapping of data to a codebook of vectors.
It was shown to be capable of modelling long term dependencies through the compressed discrete latent space, and performance was demonstrated for generating images, audio modeling, sampling conditional video sequences.
The generation of high-fidelity images was shown in \cite{razavi2019generating}, which proposed improvements the autoencoder setup from \cite{van2017neural}.

More recently, discrete representations have shown great promise in speech recognition, by enabling a differentiable mapping of spans of audio waveforms to a codebook. 
Unsupervised speech representations were learned using Wavenet autoencoders in \cite{chorowski2019unsupervised}, which also demonstrated the correspondence between phonemes and the learned symbols. 
VQ-Wav2vec \cite{baevski2019vq} pairs a future timestep prediction task with vector quantization, and studies the effectiviness of both the K-means approach from \cite{van2017neural} as well as the gumbel softmax operation \cite{gumbel1954statistical, jang2016categorical, maddison2014sampling}. 
\edit{
	Subsequently, the discrete symbols are used for RoBERTa \cite{liu2019roberta} pre-training, and the resulting embeddings are utilized by an acoustic model for improved speech recognition.
}
Using these discrete representation models now represents the state-of-the-art, with the introduction of extensions to VQ-Wav2vec, including Wav2vec2.0 \cite{baevski2020wav2vec}, unsupervised speech recognition \cite{baevski2021unsupervised}, and VQ-APC \cite{chung2020vector}.
Other works using vector quantization include w2v-BERT \cite{chung2021w2v} which sets up a masked language modeling task, and HuBERT, which also does masked prediction of hidden units \cite{hsu2021hubert}. 
\edit{
	However, these methods typically require large amounts of unlabeled data for pre-training (e.g., 960 hours of audio for a base model and 60k hours for a large model).
	Further, a language model is utilized with beam search to decode the outputs of the acoustic model.
}
Interestingly, the discrete representations enable the unsupervised discovery of acoustic units where phonemes are automatically mapped to a small set of discrete representations, enabling phoneme discovery and segmentation \cite{van2022temporal, kamper2022word, cuervo2022variable, dieleman2021variable}.
This resulting property of automatic discovery of ground truth phonemes is of particular interest, as we hypothesize that it allows us to derive the atomic units human movements from wearable sensor data, by learning a mapping of discrete representations to spans of sensor data.
We hypothesize that these movement units enable more accurate classification of activities, even with the loss of resolution due to discretization.

\subsection{Self-Supervised Representation Learning for Human Activity Recognition}
Going beyond the conventional unsupervised learning methods comprising Restricted Boltzmann Machines (RBM)s and Autoencoders \cite{haresamudram2019role, varamin2018deep}, recent years have seen the development of `self-supervised learning', that also utilize unlabeled data for representation learning. 
These methods form the `pretrain-then-finetune' training paradigm, and have resulted in significant performance improvements over end-to-end training techniques such as DeepConvLSTM \cite{ordonez2016deep}. 
The core idea is to design pretext tasks that aim at capturing specific aspects of the input, thereby resulting in useful representations for downstream recognition tasks.

Multi-task self-supervision introduced self-supervised learning to wearables-based activity recognition by performing transformation discrimination in a multi-task setting \cite{saeed2019multi}.
Eight accelerometer transformations are applied with a probability of 50\% and the network is trained to independently classify whether the transformations were applied or not. 
Subsequently, SelfHAR combined self-training with transformation discrimination by applying knowledge distillation to train a teacher network with labeled data. 
The teacher is then used to pseudo-label the unlabeled data, following which the confident samples are combined with the labeled dataset for transformation discrimination. 

Transformers were explored for self-supervision in \cite{haresamudram2020masked}, by training to reconstruct only randomly masked timesteps of windows of sensor data from mobile phones.
Contrastive Predictive Coding (CPC) was adopted and applied to wearable sensor data in \cite{haresamudram2021contrastive}, where future timestep prediction was performed under contrastive learning settings. 
A Gated Recurrent Unit (GRU) network is used to summarize sensor data encoded from a convolutional encoder and used to predict $k$ future timesteps, and optimized using the InfoNCE loss. 
Predicting multiple future timesteps captures the slowly varying signal of the data and therefore results in effective representations. 

Siamese contrastive learning using the SimCLR framework \cite{chen2020simple} was explored in \cite{tang2020exploring}.
The input windows are randomly augmented in two different ways and comprise the positive pairs, whereas the remaining pairs are the negative pairs. 
SimSiam \cite{chen2021exploring} also has a siamese setup, albeit is not trained with contrastive learning. 
The windows are augmented twice and passed through a common encoder, and one arm has an MLP whereas the other has a stop gradient operation, and is trained using a cosine distance based symmetric loss. 
BYOL \cite{grill2020bootstrap} on the other hand utilizes two networks to interact and learn from one another.
The online network learns to predict the outputs of the target while using an augmented version of the input. 
There is an asymmetry introduced in the architecture, as the target does not contain a prediction head, and mean squared error between the target representations and the normalized predictions is used to update the parameters. 
\cite{haresamudram2022assessing} studies the aforementioned methods and performs an assessment of the state-of-the-field of self-supervised human activity recognition by evaluating them on a collection of tasks, in order to understand their strengths and shortcomings.
Similarly, \cite{qian2022makes} explores these contrastive learning tasks and studies suitable augmentations and architectures for effective performance.

Enhancements to the CPC framework for wearables were investigated in \cite{haresamudram2023investigating}, by considering three components: the encoder architecture, the autoregressive network, and the future timestep prediction task. 
The modifications include: increasing the striding of the encoder, replacing the GRU with a causal convolutional network, and performing the future timestep prediction at each timestep.
The resulting `Enhanced CPC' demonstrates substantial improvements over the original framework \cite{haresamudram2021contrastive} as well as outperforms state-of-the-art self-supervision on four of six target datasets.
This superior performance, coupled with the fully convolutional architecture (which improves the parallelizability), motivates the use of Enhanced CPC as the base for discretization.

For all of the methods detailed above, the self-supervision results in dense (continuous-valued), high-dimensional representations of windows of data.
In contrast, we propose to perform \emph{discrete} representation learning, as it allows us to derive a collection of symbolic representations, which also aids in the lower-level analysis of human movements while also performing comparably to state-of-the-art self-supervision.

\section{Methodology}
\label{sec:method}
\begin{figure}[!t]
	\centering
	\includegraphics[height=3.0cm]{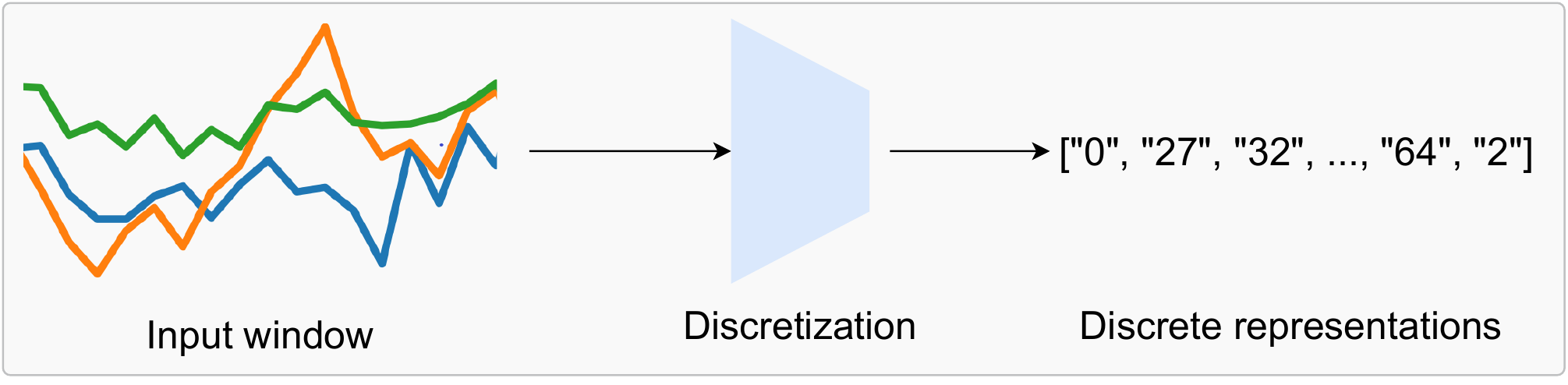}
	\caption{
		\edit{
		Discrete representation learning: given a window of accelerometer data as input, the output is a collection of enumerated symbols.
		As such, we map short spans of continuous-valued time-series sensor data to a list of `symbols' (i.e., the numbers in the figure). 
		Therefore, we obtain the ``strings of motion'', as the symbols are discrete.
	}
	}
	\label{fig:discretization_overview}
\end{figure}

\begin{figure}[!t]
	\centering
	\includegraphics[width=\textwidth]{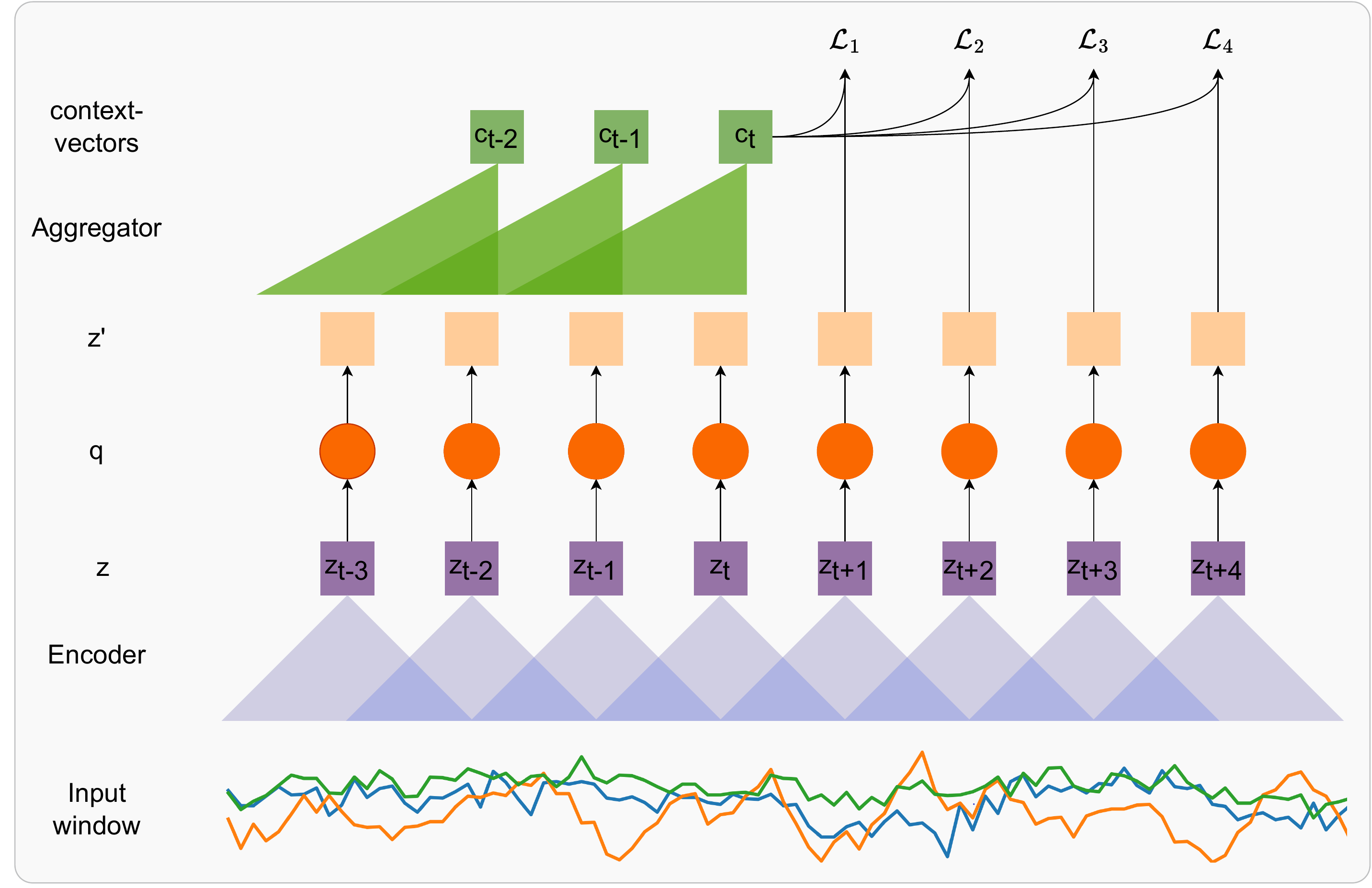}
	\caption{
		Overview of the discrete representation learning framework for human activity recognition from wearables. 
		We combine a contrastive future timestep prediction problem with vector quantization to map spans of sensor data to a codebook of vectors.
		The index of the codebook vector closest to the latent representation $z_t$ functions as the discrete representation.
	}
	\label{fig:vq_cpc}
\end{figure}

In this paper, we introduce the discrete representation learning framework for wearable sensor data, with the ultimate goal of improved activity recognition performance and better analysis of human movements.
This paper represents the first step towards this goal, which--effectively--demonstrates the proof-of-concept for the effectiveness of learned discretization, which warrants the aforementioned ``return to discretized representations''.
Based on our framework, we explore the potential and next steps for discretized human activity recognition.
\edit{
	An overview of discretization is shown in Fig.\ \ref{fig:discretization_overview}, which involves mapping windows of time-series accelerometer data to a collection of discrete `symbols' (which are represented by strings of numbers).
}

Following the self-supervised learning paradigm, our approach contains two stages: 
\emph{(i)} pre-training, where the network learns to map unlabeled data to a codebook of vectors, resulting in the discrete representations; and 
\emph{(ii)} fine-tuning, which utilizes the discrete representations as input for recognizing activities.
In order to enable the mapping, we apply Vector Quantization (VQ) to the Enhanced CPC framework.
While VQ can be applied for end-to-end training as well, self-supervision enables us to derive the discrete representations in an unsupervised manner, thereby leveraging the availability of large-scale in-the-wild datasets.
Therefore, the base of the discretization process is self-supervision, where the loss from the pretext task is added to the loss from the VQ module in order to update the network parameters as well as the codebook vectors.

To this end, we first detail the self-supervised pretext task, Enhanced CPC, and describe how the VQ module can be added to it. 
With the aim of quantitatively measuring the utility of the representations, we perform activity recognition with the discrete representations derived from downstream target labeled datasets. 
Therefore, we also discuss the classifier network used for such evaluation and clarify how the setup is different from state-of-the-art self-supervision for wearables.

\subsection{Discrete Representation Learning Setup}
\label{sec:disrete_rep_learning_setup}
The setup for learning discrete representations of human movements contains two parts: 
\emph{i)} the self-supervised pretext task; and 
\emph{ii)} the Vector Quantization (VQ) module.
We utilize the Enhanced Contrastive Predictive Coding (CPC) framework \cite{haresamudram2023investigating} as the self-supervised base, which comprises the prediction of multiple future timesteps in a contrastive learning setup. 
By predicting farther into the future, the network can capture the slowly varying features, or the long-term signal present in the sensor data while ignoring local noises, which is beneficial for representation learning \cite{oord2018representation}.
We borrow notation from \cite{baevski2019vq} for the method description below. 

The aim of the Enhanced CPC \cite{haresamudram2023investigating} framework is to investigate three modifications to the original wearables-based CPC framework: \emph{(i)} the convolutional encoder network; \emph{(ii)} the Aggregator (or autoregressive network); and \emph{(iii)} the future timestep prediction task.
First, the encoder from \cite{haresamudram2021contrastive} is replaced with a network with higher striding (details below), resulting in a reduction in the temporal resolution.
In addition, a causal convolutional network is used to summarize previous latent representations into a context vector instead of the GRU-based autoregressive network.
Finally, the future timestep prediction is performed at every context vector instead of utilizing a random timestep to make the prediction.
These changes, put together, substantially improve the performance of the learned Enhanced CPC representations, compared to state-of-the-art methods.
In what follows, we provide the architectural details and a detailed description of the technique.

As shown in Fig.\ \ref{fig:vq_cpc}, we utilize a convolutional encoder to map windows of sensor data to latent representations $f:\mathcal{X} \mapsto \mathcal{Z}$ (called $z-$vectors).
The conv. encoder comprises four blocks, each containing a 1D convolutional network followed by the ReLU activation and dropout with p=0.2. 
The layers consist of (32, 64, 128, 256) channels respectively, with a kernel size of (4, 1, 1, 1) and a stride of (2, 1, 1, 1).
\edit{
	The encoder output frequency is 24.5 Hz, as we obtain 49 $z-$vectors for each window of 100 timesteps (i.e., two seconds of data at 50 Hz). 
	Therefore, we obtain one $z_t$ for approx.\ every \emph{two} timesteps of data.
	By adjusting the convolutional encoder architecture appropriately, the frequency can be adjusted to increase or reduce relative to the base setup detailed above (see Sec.\ \ref{sec:symbol_duration}). 
}
\edit{
	In addition, the convolutional encoder can also be modified for training on data recorded at higher sampling rates (i.e., $>50$ Hz), in order to maintain an output frequency of $z-$vectors at 24.5 Hz. 
}

The quantization module ($q:\mathcal{Z} \mapsto \hat{\mathcal{Z}}$) replaces each $z_t$ with $\hat{z}=e_i$, which is the index of the closest codebook vector (also called the codeword), from a fixed size codebook $e \in \mathbb{R}^{V \times d}$, containing $V$ representations of size $d$ (details in Sec.\ \ref{sec:kmeans_quantization}).
We utilize the online K-means based quantization from \cite{baevski2020wav2vec}, which is similar to the vector quantized autoencoder \cite{van2017neural} detailed originally in \cite{van2017neural}. 

Following the Enhanced CPC framework,
a causal convolutional network called the `Aggregator' is used for summarizing previous timesteps of encoded representations $\hat{z}_{\leq t}$ ($g: \hat{\mathcal{Z}} \mapsto \mathcal{C}$) into the context vectors $c_t$, which are used to predict multiple future timesteps.
This enables improved parallelization due to the convolutions and results in faster training times.
Each block in the Aggregator has 256 filters with dropout p=0.2, layer normalization, and residual connections between layers, as utilized in \cite{baevski2020wav2vec}.
For each causal convolution layer in successive blocks, the stride is set to 1 whereas the kernel sizes are consecutively increased from 2.
The network is once again trained to identify the ground truth $z_{t+k}$, which is $k$ steps in the future from a collection of negatives sampled randomly from the batch, for every $c_t$ in the window.
Such a setup was first introduced in VQ-Wav2vec \cite{baevski2020wav2vec}, where two quantization approaches--Gumbel softmax \cite{gumbel1954statistical} and K-means \cite{van2017neural, baevski2020wav2vec}--were studied for their effectiveness towards better speech recognition. 
In our work however, preliminary explorations revealed the higher effectiveness of the online K-means-based quantization, described below.

\subsubsection{K-means Quantization}
\label{sec:kmeans_quantization}

\begin{figure}
	\centering
	\includegraphics[width=0.5\textwidth]{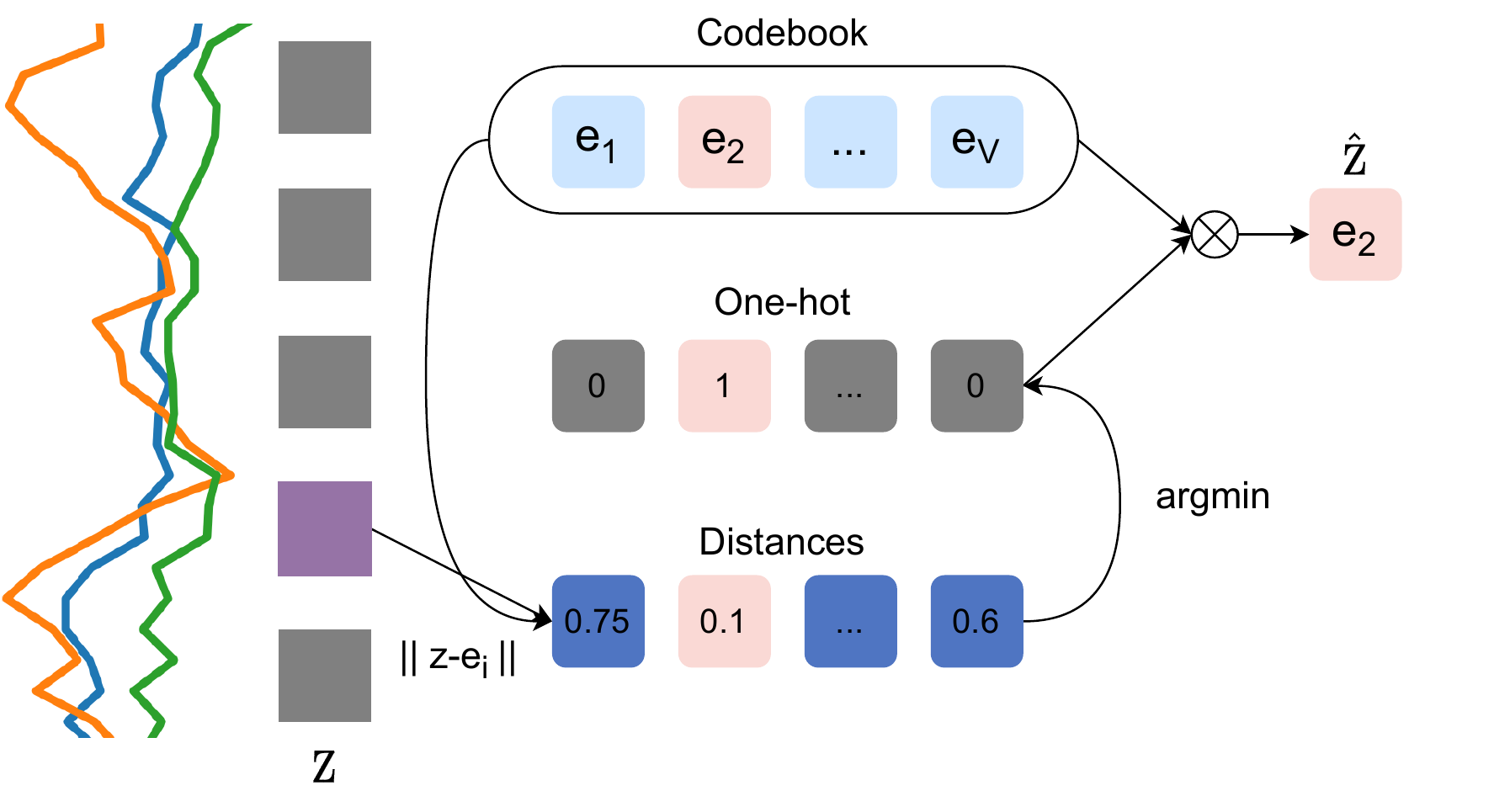}
	\caption{
		\edit{
		Visualizing Kmeans based quantization: for each $z-$vector, the $l_2$ distance is computed to the codebook of vectors $e_i$.
		The index of the nearest codebook vector comprises the discrete representation, whereas the nearest vector itself is passed as the output (i.e., $\hat{z}-$vector).
		This figure has been adopted from \cite{baevski2019vq}.
		} 
	}
	\label{fig:kmeans_quant}
\end{figure}

As detailed previously, the codebook has a size of $V \times d$, where $V$ is the number of variables in the codebook, and $d$ is their dimensionality.
The vector quantization procedure allows for a differentiable process to select codebook indices.
\edit{
As shown in Fig.\ \ref{fig:kmeans_quant}, 
} the nearest neighbor codebook vector to any $z-$vector in terms of the Euclidean distance is chosen, yielding $i = argmin_{j} \Vert z-e_j \Vert^2_2$.
The $z-$vector is replaced $\hat{z}=e_i$, which is the codebook vector at the index $i$.
As mentioned in \cite{van2017neural}, this process can be considered a non-linearity that maps latent representations to one of the codebook vectors.

As choosing the codebook indices does not have a gradient associated with it, the straight-through estimator \cite{bengio2013estimating} is employed to simply copy gradients from the Aggregator input $q(z)$ to the encoder output $f(x)$.
Therefore, the forward pass comprises the selection of the closest codebook vector, whereas during the backward pass, the gradient gets copied as-is to the encoder.
The parameters are updated using the future timestep prediction loss as well as two additional terms:

\begin{equation}
	\mathcal{L} = \sum_{k=1}^{K} \mathcal{L}^{CPC}_k + \big( \Vert sg(z) - \hat{z} \Vert ^2 + \gamma \Vert z - sg(\hat{z}) \Vert ^2 \big)
\end{equation}

\noindent where $sg(x) \equiv x$, $\frac{d}{dx}sg(x) \equiv 0$ is the stop gradient operator, $k$ is the future timestep, and $\gamma$ is a hyperparameter. 
Due to the straight-through estimation, the codebook does not obtain any gradients from $\mathcal{L}^{CPC}$.
However, the second term $\Vert sg(z) - \hat{z} \Vert ^2$ moves the codebook vectors closer to the $z-$vectors, whereas the third term $\Vert z - sg(\hat{z}) \Vert ^2$ ensures that $z-$vectors are close to a codeword.
Therefore, the Aggegator network is updated via the first loss term, whereas the convolutional encoder is optimized by the first and third loss terms, and the codebook vectors are updated using the second loss term.
\edit{
	This is  visualized in Fig.\ \ref{fig:vq_cpc_losses} in the Appendix. 
}
The weighting term $\gamma$ is set to $0.25$ as utilized in \cite{van2017neural, baevski2019vq}, as we obtained good performance.

\subsubsection{Preventing Mode Collapse}
\label{sec:prev_mode_collapse}
As discussed in \cite{baevski2020wav2vec}, replacing $z$ by a single entry $e_i$ from the codebook is prone to mode collapse, where very few (or only one) codebook vectors are actually used.
This leads to very poor outcomes due to a lack of diversity in the discrete representations.
To mitigate this issue, \cite{baevski2020wav2vec} suggests independent quantization of partitions, such that $z \in \mathbb{R}^{d}$ is organized into multiple groups $G$ using the form $z \in \mathbb{R}^{G \times (\frac{d}{G})}$.
Each row is represented by an integer index, and the discrete representation is given by indices $i \in [V]^G$, where $V$ is the number of codebook variables for the particular group and each element $i_j$ is the index of a codebook vector.
For each of the groups, the vector quantization is applied and the codebook weights are not shared between them.
During pre-training, we utilize $G=2$ (as per \cite{baevski2020wav2vec}), and $V=100$, resulting in a possible $V^G$ possible codewords.
In practice, the number of unique discrete representations is generally significantly smaller than  $100^2$.

\subsection{Classifier Network}
\label{sec:classifier_network}

\begin{figure}[!t]
	\centering
	\includegraphics[height=3.0cm]{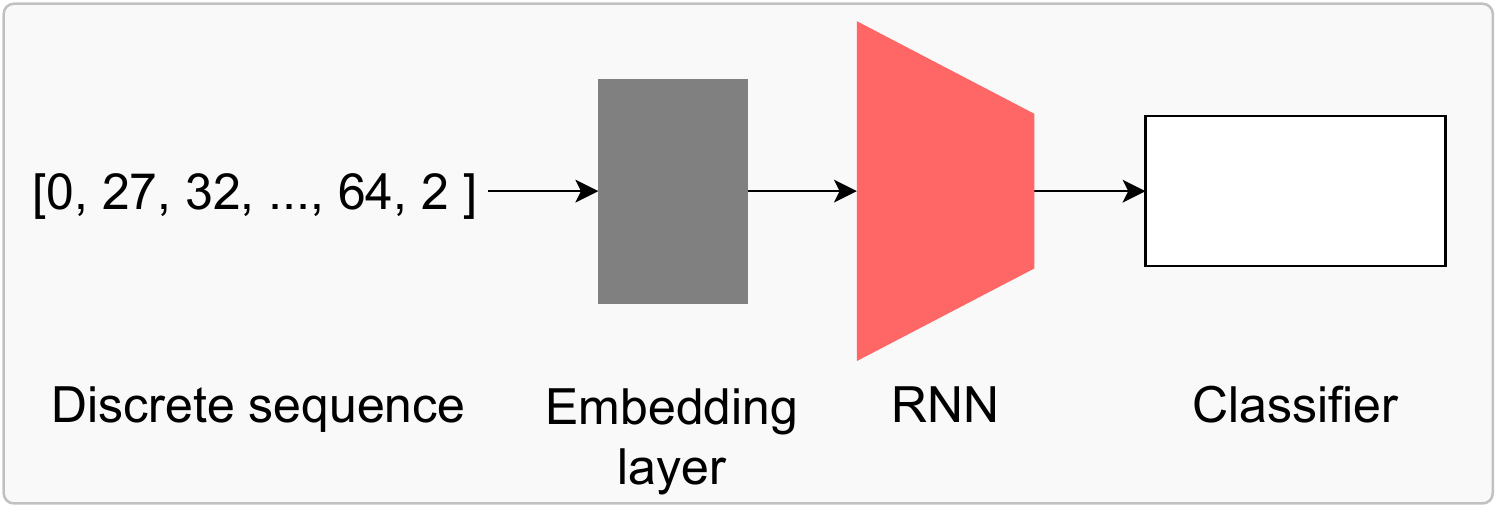}
	\caption{
		\edit{
		Performing classification using the discrete representations: the sequences of symbolic representations (indexed) are first passed through a learnable embedding layer.
		Subsequently, an RNN network (GRU or LSTM) is utilized along with an MLP network for classifying the sequences into the activities of interest.
		}
	}
	\label{fig:classification}
\end{figure}

As the obtained representations (or symbols) are discrete in nature, applying a classifier directly is not possible.
Therefore, we utilize an established setup from the natural language processing domain (which also deals with discrete sequences) to perform activity recognition, shown in Fig.\ \ref{fig:classification}. 

First, the discrete representations are indexed (assigned a number based on the total number of such symbols present in the data).
For each window of symbols, we append the $\langle START \rangle$ and $\langle END \rangle$ tokens to the beginning and end of the window.
The dictionary also contains the pad $\langle PAD \rangle$ and unknown $\langle UNK \rangle$ tokens, which represent padding (sequences of differing lengths can be padded to a common length) and unknown (symbols present during validation/test but not during training, for example).
The indexed sequences are used as input to a learnable embedding layer (shown in grey in Fig.\ \ref{fig:classification}), followed by a LSTM or GRU network of 128 nodes and two layers with dropout with p=0.2.
Subsequently, a MLP network identical to the classifier network from \cite{haresamudram2022assessing} is applied.
It contains three linear layers of 256, 128, and $num\_classes$ units with batch normalization, ReLU activation and dropout in between.

\section{Setup}
\label{sec:settings}
In Sec.\ \ref{sec:method}, we introduced an overview of our framework for deriving discrete representations from sensor data and for performing a quantitive evaluation using activity recognition.
Here, we describe the setup utilized to learn such representations, including the datasets utilized for pre-training and evaluation (Sec.\ \ref{sec:datasets}), the data pre-processing (Sec.\ \ref{sec:data_preparation}), and the implementation details (Sec.\ \ref{sec:implementation_details}).
Put together, these provide an overview of the practical details vital for learning and utilizing discrete representations for sensor-based HAR.

\subsection{Datasets}
\label{sec:datasets}
Both the pre-training and the classification are performed using data from a single accelerometer, as we can reasonably expect a single wearable to be feasible in most scenarios. 
Pre-training is performed using the Capture-24 dataset, which contains a single wrist-worn accelerometer data.
\edit{
	We chose Capture-24 primarily due to its large-scale and recording setup: it contains around 2,500 hours of data from 151 participants in daily living conditions (i.e., in-the-wild), thereby not limiting the types of movements and activities recorded (see Sec. \ref{sec:capture_24} for details). 
	In addition, prior works such as \cite{haresamudram2022assessing, haresamudram2023investigating} have utilized it as the base for self-supervised pre-training, allowing us to compare our results against those works.
}
Based on the assessment framework in \cite{haresamudram2022assessing}, the performance of the discrete representations is evaluted on target datasets collected at the wrist, waist, and the leg, albeit we utilize two datasets per location (unlike \cite{haresamudram2022assessing}, which uses three).
The source (Capture-24) and target datasets are summarized in Tab.\ \ref{tab:datasets}, and briefly discussed below.
As in \cite{haresamudram2022assessing}, we downsample all datasets to 50 Hz.

\begin{table*}[!t]
	\centering
	\small
	\caption{
		Summary of the datasets used in our study.
		Capture-24 is the source dataset (wrist) whereas the others comprise the target, spread across three sensor locations -- wrist, waist, and the leg/ankle (adopted / adapted with permission from  \cite{haresamudram2022assessing}).
	}
	\begin{tabular}{P{2.2cm} P{1.5cm} c c P{8.0cm}}
		\toprule
		Dataset & Location & \# Users & \# Act. & Activities \\ 
		\midrule
		Capture-24 \cite{willetts2018statistical, chan2021capture, gershuny2020testing} & Wrist & 151 & N/A & Free living \\
		\hdashline\noalign{\vskip 0.5ex}
		
		HHAR \cite{stisen2015smart} & Wrist & 9 & 6 & Biking, sitting, going up and down the stairs, standing, and walking \\ 
		Myogym \cite{koskimaki2017myogym} & Wrist & 10 & 31 & Seated cable rows, one-arm dumbbell row, wide-grip pulldown behind the neck, bent over barbell row, reverse grip bent-over row, wide-grip front pulldown, bench press, incline dumbbell flyes, incline dumbbell press and flyes, pushups, leverage chest press , close-grip barbell bench press, bar skullcrusher, triceps pushdown, bench dip, overhead triceps extension, tricep dumbbell kickback, spider curl, dumbbell alternate bicep curl, incline hammer curl, concentration curl, cable curl, hammer curl, upright barbell row, side lateral raise, front dumbbell raise, seated dumbbell shoulder press, car drivers, lying rear delt raise, null \\ 
		\hdashline\noalign{\vskip 0.5ex}
		
		Mobiact \cite{chatzaki2016human} & Waist/ Trousers & 61 & 11 & Standing, walking, jogging, jumping, stairs up, stairs down, stand to sit, sitting on a chair, sit to stand, car step-in, and car step-out\\ 
		Motionsense \cite{malekzadeh2018protecting} & Waist/ Trousers & 24 & 6 & Walking, jogging, going up and down the stairs, sitting and standing \\ 
		\hdashline\noalign{\vskip 0.5ex}
		
		MHEALTH \cite{banos2014mhealthdroid, banos2015design} & Leg/ Ankle & 10 & 13 & Standing, sitting, lying down, walking, climbing up the stairs, waist bend forward, frontal elevation of arms, knees bending, cycling, jogging, running, jump front and back\\ 
		PAMAP2 \cite{reiss2012introducing} & Leg/ Ankle & 9 & 12 & Lying, sitting, standing, walking, running, cycling, nordic walking, ascending and descending stairs, vaccuum cleaning, ironing, rope jumping \\ 
		\bottomrule
	\end{tabular}
	\label{tab:datasets}
\end{table*}

\subsubsection{Capture-24}
\label{sec:capture_24}
It is a large-scale dataset recorded from a single wrist-worn accelerometer based on the Axivity AX3 platform \cite{chan2021capture, willetts2018statistical, gershuny2020testing}.
It comprises free living conditions, with 151 users of data for approximately one day each, resulting in around 4,000 hours of data in total. 
Out of this, 2,500 hours are coarsely labeled into the six broad activities  sleep, sit-stand, mixed, walking, vehicle, and bicycling.
There are also 200 fine-grained activities, and the annotation was performed using a chest-mounted Vicon Autograph and Whitehall II sleep diaries.
Going by the broad labels, around 75\% of the data is either sleep or sit-stand, thereby rendering it imbalanced.

\subsubsection{HHAR}
The underlying goal for the collection of this dataset was to study the impact of heterogenous recording devices (incl.\ sensors, devices, and workloads) on recognition of human activities \cite{stisen2015smart}.
It contains data from both mobile phones (LG Nexus 4, Samsung Galaxy S Plus, Samsung Galaxy S3, Samsung Galaxy S3 mini) and smartwatches (LG G and Samsung Galaxy Gear).
We only utilize the data from the wrist watches, which were worn on each arm, as we are interested in studying the performance obtained at the wrist.
Data was collected from nine users in total, who performed five minutes per activity, resulting in a balanced dataset.

\subsubsection{Myogym}
The Myogym \cite{koskimaki2017myogym} dataset was collected from 10 participants performing thirty different gym activities (or the NULL class), where each activity contains ten repititions.
The Myo armband on the right forearm was utilized for recording the data, and comprises an IMU containing an accelerometer, gyroscope, and magnetometer, along with eight electromyogram (EMG) sensors. 
Data were recorded at 50 Hz and the aim of the study was to faciliate activity and gesture recognition, as well as sensor fusion. 

\subsubsection{Mobiact}
A Samsung Galaxy S3 smartphone placed freely in a trouser pocket was utilized to four types and twelve locomotion-style + transitionary activities \cite{chatzaki2016human} at a rate of 200 Hz.
Following \cite{saeed2019multi}, we removed the lying class, resulting in eleven activities from a total of 61 participants (out of 66).
The sensors include accelerometer, gyroscope, and orientation, and we utilize v2 of this dataset (as in \cite{saeed2019multi}).

\subsubsection{Motionsense}
It contains data recorded from an iPhone 6s, including accelerometer, gyroscope, and attitude information at a rate of 50 Hz \cite{malekzadeh2018protecting}.
A total of 24 subjects (14 men and 10 women) were recruited for data collection, with the aim of developing privacy preserving sensor data transmission systems.
It mainly contains locomotion-style activities (see \autoref{tab:datasets}), collected from the front trouser pocket.

\subsubsection{MHEALTH}
The MHEALTH dataset \cite{banos2014mhealthdroid, banos2015design} consists of data from 10 participants performing 12 activities. 
Shimmer 2 \cite{burns2010shimmer} wearable sensors were utilized for the recording, and placed on the chest, right wrist, and left ankle.
The sampling rate is 50 Hz and the activities under study include locomotion along with some exercises. 
The collection was performed out of the laboratory, without any constraints on how they had to be executed; the subjects were asked to try their best while executing the activities.

\subsubsection{PAMAP2}
This dataset comprises three IMUs and a heart rate (HR) monitor, recorded to facilitate the development of physical activity monitoring systems \cite{reiss2012introducing}.
One IMU is placed at the chest along with the HR monitor, whereas the remaining two are placed on the dominant wrist and the ankle.
A total of 9 participants (8 males + 1 female) are preset in the dataset and followed a protocol of 12 activities (listed in \autoref{tab:datasets}) along with (watch TV, computer work, drive car, fold laundry, clean house, and play soccer) optionally. 
In our study, we only utilize the 12 that form a part of the protocol for collection.
Further, we only utilize data from the ankle-worn accelerometer, as we evaluate the performance across sensor locations as well.

\subsection{Data Pre-Processing}
\label{sec:data_preparation}
Our study utilizes data from a single accelerometer for both pre-training and evaluation. 
The source dataset is collected at the wrist, whereas the target datasets comprise the wrist, waist, and leg locations.
For pre-training, the sampling rate of Capture-24 is reduced to 50 Hz \edit{by sub-sampling} so as to reduce the computational load and training times (identical to \cite{haresamudram2022assessing}).
We also downsample all target datasets to 50 Hz \edit{via sub-sampling}  (if they were higher originally), as it was shown in \cite{haresamudram2022assessing}, matching the sampling rates between pre-training and fine-tuning is important for optimal performance.

Following \cite{haresamudram2022assessing}, the window size is set to 2 seconds with an overlap of 0 (for Capture-24) and 50\% (for target datasets), in order to ensure that both long- and short-term activities are sufficiently captured in any randomly picked window.
For Capture-24, the dataset is split randomly by participants at a 90:10 ratio for training and validation. 
The train split was normalized to have zero mean and unit variance, and the resulting means and variances were applied to the validation split as well.
Further, we only pre-train on randomly sampled 10\% of the windows from the train split, as it was shown to have comparable performance to using the entire dataset in \cite{haresamudram2022assessing}, which also reduces the time taken for pre-training.

For evaluation, the target datasets are separated into five folds: the first fold is split at a 80:20 ratio by participant into the train-val and test sets. 
The train-val set is once again partitioned randomly by participant IDs at an 80:20 ratio into the training and validation splits. 
For the remaining folds, 20\% of the participants are chosen randomly to be test set, such that no participant appears in more than one test set, across five folds. 
The train and validation splits are constructed from the remaining participants (i.e., participants that are not a part of the test set), once again at a 80:20 ratio.
The means and variances from the Capture-24 train split are also applied to all sets from the target datasets, for improved performance (as per \cite{haresamudram2022assessing}).

\subsection{Implementation Details}
\label{sec:implementation_details}
All models were implemented using the Pytorch framework \cite{paszke2019pytorch}.
For the self-supervised baselines, including Multi-task self-supervision, Autoencoder, SimCLR, CPC, and Enhanced CPC, we report the performance detailed in the original Enhanced CPC paper \cite{haresamudram2023investigating}.\footnote{We will release the code once the paper is accepted for publication.} 

For the discrete version of CPC, we set the learning rate and L2 regularization during pre-training to $1e-4$ and tune over the number of convolutional aggregator layers $\in \{ 2, 4, 6\}$.
The loss weighting parameter, $\gamma$, is set to 0.25 as recommended in \cite{van2017neural, baevski2019vq}.
In addition, we find that a prediction horizon of $k=10$ with number of negatives = 10, is sufficient for effective training. 
For most experiments (save Sec. \ref{sec:symbol_duration}), each symbol spans approx.\ two timesteps, as we found it to have a good balance between the resulting pre-training times and the temporal resolution.

The pre-training is performed for a maximum of 50 epochs, but early stopping with a patience of 5 epochs is also employed to terminate training if the validation loss does not improve.
A cosine learning rate schedule is employed -- first, the learning rate is warmed up linearly to $1e-4$ (as mentioned above) for a duration of 8\% of the total number of updates.
Subsequently, the learning rate is decayed to zero using a cosine function.
The early stopping only begins after $20$ epochs, in order to ensure the completion of warmup, and sufficient training before the termination of pre-training.
The Adam \cite{kingma2014adam} optimizer is utilized with a batch size of 128. 

The evaluation with the RNN classifier is also performed for 50 epochs, where the learning rate and L2 regularization are tuned over $\{1e-3, 1e-4, 5e-4\}$ and $\{0, 1e-4, 1e-5\}$ respectively.
Once again, the Adam optimizer is utilized, with a batch size of 256.
The learning rate is decayed by a factor of 0.8 every 10 epochs.
\edit{
	We average the validation F1-scores across the folds in order to identify the best performing hyperparameter combination.
	The corresponding average test set F1-score across the folds (for the best hyperparameters) is reported in Tab.\ \ref{tab:diff_locs}, where five randomized runs are also performed.
}
The best performing hyperparameters for GRU based evaluation are listed in Tab. \ref{tab:hyperparameters} below for easy reference.

\begin{table*}[!t]
	\centering
	\small
	\caption{
		Hyper-parameters utilized for pre-training and evaluation on the target datasets with the GRU classifier. 
	}
	\begin{tabular}{c c c c c c c c c}
		\toprule
		& \multicolumn{6}{c}{Pre-training} & \multicolumn{2}{c}{Evaluation} \\ 
				\cmidrule(lr){2-7} \cmidrule(lr){8-9}
		\multirow{-2}{*}{Dataset} & lr & l2 reg. & \# Conv. agg. & k & \# neg. & $\gamma$ & lr & wd \\ 
		\midrule
		HHAR & 1e-4 & 1e-4 & 2 & 10 & 10 & 0.25 & 5e-4 & 1e-4 \\ 
		Myogym & 1e-4 & 1e-4 & 2 & 10 & 10 & 0.25 & 5e-4 & 1e-5 \\ 
		\hdashline\noalign{\vskip 0.5ex}
		
		Mobiact & 1e-4 & 1e-4 & 2 & 10 & 10 & 0.25 & 1e-3 & 1e-4 \\ 
		Motionsense & 1e-4 & 1e-4 & 2 & 10 & 10 & 0.25 & 5e-4 & 0.0 \\ 
		\hdashline\noalign{\vskip 0.5ex}
		
		MHEALTH & 1e-4 & 1e-4 & 2 & 10 & 10 & 0.25 & 5e-4 & 1e-5 \\ 
		PAMAP2 & 1e-4 & 1e-4 & 2 & 10 & 10 & 0.25 & 1e-3 & 1e-4 \\ 
		\bottomrule
	\end{tabular}
	\label{tab:hyperparameters}
\end{table*}

\section{Results}
\label{sec:results}
Through the work presented in this paper we aim to demonstrate the potential of learning discrete representations of human movements. 
For this, we first evaluate their effectiveness for recognizing the activities performed in windows of sensor data via a simple recurrent classifier.
The performance is contrasted against established supervised baselines (such as DeepConvLSTM), as well as the state-of-the-art for representation learning, which is self-supervision.
Subsequently, we contrast the impact of \emph{learning} the discrete representations rather than computing via prior methods involving SAX.
This is followed by an exploration into the discrete reprsentation learning framework, where we study the impact of controlling the resulting alphabet size (i.e., the fidelity of the representations), and the effect of the duration of sensor data each symbol representations.
Finally, we apply self-supervised pre-training techniques designed  for discrete sequences, in order to study whether such tasks can further help improve recognition performance. 
Overall, these experiments are designed to not only study whether discrete representations can be useful, but also to derive a deeper understanding into their working.

\subsection{Activity Recognition with Discrete Representations}
\begin{table*}[!t]
	\centering
	\small
	\caption{
		Activity recognition performance: we report the mean and standard deviation of the five-fold test F1-score across five randomized runs.
		The best performing technique overall for each dataset is denoted in \textbf{\darkgreen{bold}}, whereas the best unsupervised method is shown using $^\dagger$.
		Therefore, methods with \textbf{\darkgreen{bold}}$^\dagger$ are the best method overall and the best unsupervised method as well.
		The performance for the methods with $^*$ was obtained from \cite{haresamudram2023investigating}.
		We observe that discrete representations show comparable if not better performance to self-supervision on three of the benchmark datasets, indicating the capabilities of the learned symbols.
	}
	\begin{tabular}{ccccccc}
		\toprule
		& \multicolumn{2}{c}{Wrist} & \multicolumn{2}{c}{Waist} & \multicolumn{2}{c}{Leg} \\ 
		\multirow{-2}{*}{Method} & HHAR & Myogym & Mobiact & Motionsense & MHEALTH & PAMAP2 \\ 
		\midrule
		\rowcollight \multicolumn{7}{c}{Supervised baselines} \\ 
		Conv. classifier$^*$ & 55.63 $\pm$ 2.05 & 38.21 $\pm$ 0.62 & 78.99 $\pm$ 0.38 & 89.01 $\pm$ 0.89 & 48.71 $\pm$ 2.11 & 59.43 $\pm$ 1.56 \\ 
		DeepConvLSTM$^*$ & 52.37 $\pm$ 2.69 & 39.36 $\pm$ 1.56 & \textbf{\darkgreen{82.36 $\pm$ 0.42}} & 84.44 $\pm$ 0.44 & 44.43 $\pm$ 0.95 & 48.53 $\pm$ 0.98 \\ 
		GRU classifier$^*$ & 45.23 $\pm$ 1.52 & 36.38 $\pm$ 0.60 & 75.74 $\pm$ 0.60 & 87.42 $\pm$ 0.52 & 44.78 $\pm$ 0.47 & 54.35 $\pm$ 1.64 \\
		%
		\rowcollight \multicolumn{7}{c}{Self-supervision + MLP classifier} \\ 
		Multi-task self.\ sup$^*$ & 57.55 $\pm$ 0.75 & 42.73 $\pm$ 0.49 & 72.17 $\pm$ 0.38 & 86.15 $\pm$ 0.42 & 50.39 $\pm$ 0.72 & \textbf{\darkgreen{60.25 $\pm$ 0.72}}$^\dagger$ \\ 
		Autoencoder$^*$ & 53.64 $\pm$ 1.04 & 46.91 $\pm$ 1.07 & 72.19 $\pm$ 0.35 & 83.10 $\pm$ 0.60 & 40.33 $\pm$ 0.37 & 59.69 $\pm$ 0.72 \\ 
		SimCLR$^*$ & 56.34 $\pm$ 1.28 & \textbf{\darkgreen{47.82 $\pm$ 1.03}}$^\dagger$ & 75.78 $\pm$ 0.37 & 87.93 $\pm$ 0.61 & 42.11 $\pm$ 0.28 & 58.38 $\pm$ 0.44 \\ 
		CPC$^*$ & 55.59 $\pm$ 1.40 & 41.03 $\pm$ 0.52 & 73.44 $\pm$ 0.36 & 84.08 $\pm$ 0.59 & 41.03 $\pm$ 0.52 & 55.22 $\pm$ 0.92 \\ 
		Enhanced CPC$^*$ & 59.25 $\pm$ 1.31 & 40.87 $\pm$ 0.50 & 78.07 $\pm$ 0.27$^\dagger$ & \textbf{\darkgreen{89.35 $\pm$ 0.32}}$^\dagger$ & \textbf{\darkgreen{53.79 $\pm$ 0.83}}$^\dagger$ & 58.19 $\pm$ 1.22 \\ 
		%
		\rowcollight \multicolumn{7}{c}{Discrete representations + RNN classifier} \\ 
		VQ CPC + LSTM class.  & \textbf{\darkgreen{60.76 $\pm$ 1.09}}$^\dagger$ & 29.62 $\pm$ 0.52 & 76.34 $\pm$ 0.30 & 89.06 $\pm$ 0.24 & 48.86 $\pm$ 0.34 & 55.28 $\pm$ 0.34 \\ 
		VQ CPC + GRU class.  & 60.26 $\pm$ 0.83 & 31.65 $\pm$ 0.29 & 77.78 $\pm$ 0.17 & 89.23 $\pm$ 0.23 & 49.01 $\pm$ 0.30 & 56.92 $\pm$ 0.26 \\ 
		\bottomrule
	\end{tabular}
	
	\label{tab:diff_locs}
\end{table*}

First, we compare the performance of the discrete representations towards recognizing activities from windows of discrete sequential data.
Once the pre-training is complete, we perform inference to obtain the discrete representations and utilize the setup detailed in Sec.\ \ref{sec:classifier_network} for classification.
The performance is compared against diverse self-supervised learning techniques, which form the state-of-the-art for representation learning in HAR
(as shown in \cite{haresamudram2022assessing}), including: 
\emph{i)} Multi-task self-supervision, which utilizes transformation discrimination as the pretext task; 
\emph{ii)} Autoencoder, reconstructing the original input through an additional decoder network; 
\emph{iii)} SimCLR, contrasting two augmented versions of the same input window against negative pairs from the batch; 
\emph{iv)} CPC, which uses multiple future timestep predicton for pre-training; and 
\emph{v)} Enhanced CPC, as before but with improvements to the CPC framework on the encoder, aggregator, and future prediction tasks \cite{haresamudram2023investigating}.
For these techniques, the encoder weights are frozen and only the MLP classifier is updated with label information. 

DeepConvLSTM, a Conv. classifier with the same architecture as the encoder for Multi-task, Autoencoder, and SimCLR, along with a GRU classifier function as the end-to-end training baselines. 
We perform five fold cross validation and report the performance for five randomized runs in Tab.\ \ref{tab:diff_locs}. 
The comparison is performed on six datasets across sensor locations (Capture-24 is collected at the wrist, whereas the target datasets are spread across the wrist, waist, and the leg) and activities (which include locomotion, daily living, health exercises, and fine-grained gym exercises).

For the waist-based Mobiact, which covers locomotion-style activities along with transitionary classes such as stepping in and out of a car, the discrete representation learning performs comparably or better than all methods, obtaining a mean of 77.8\%.
However, for Motionsense, the performance is similar to the best performing model overall, which is Enhanced CPC, once again outperforming other self-supervised and supervised baselines.
Considering the leg-based PAMAP2 dataset, VQ-CPC obtains lower performance and is similar to the GRU classifier. 
For MHEALTH as well, the performance drops significantly compared to Enhanced CPC, showing a reduction of around 4.8\%, yet outperforming the Autoencoder, SimCLR, Multi-task, and CPC.

Finally, we consider the wrist-based datasets such as HHAR and Myogym.
HHAR comprises locomotion-style activities and the discrete representations  improve the performance over Enhanced CPC, by around 1.5\%, thereby constituting the best option for wrist-based recognition of locomotion activities.
Interestingly, the discretization results in poor features for classifying fine-grained gym activities, with the performance dropping significantly compared to other self-supervised methods. 
Enhanced CPC also sees substantially lower performance than SimCLR, likely due to the increased striding in the encoder, which results in a latent representation for approx.\ every second timestep, thereby negatively impacting the recognition of activities such as fine-grained curls and pulls. 
In addition, the discretization results in a smaller, finite codebook, which is a loss in temporal resolution compared to continuous-valued high-dimensional features. 
This is detrimental for Myogym, resulting in the poor performance.

Therefore, the discrete representations can result in effective recognition of locomotion-style and daily living activities, and overall perform the best (or similar to the best) on three benchmark datasets, at the wrist and waist.
The loss in resolution due to mapping the continuous-valed sensor data to a finite collection of codebook vectors (and their indices) does not have a significant negative impact on locomotion-style activities, but is detrimental for recognizing fine-grained movements (as present in Myogym for example).
In addition, the effective performance across sensor location indicates the representation learning capabilities of the discrete representation learning process, and shows its promise for sensor-based HAR.
This result presents practioners with a new option for activity recognition, with comparable performance and potentially lowered data upload costs, as the discretized representations result in more compressed data than continuous-valued sensor readings.

\subsection{Comparison to Established Discretization Methods}
\begin{table*}[!t]
	\centering
	\small
	\caption{
		Comparing the performance of the proposed discrete representation learning technique to SAX and SAX-REPEAT (multi-variate version of SAX):  
		the computed symbolic representations are evaluated on identical LSTM classifiers described in Sec. \ref{sec:classifier_network}. 
		SAX and SAX-REPEAT perform poorly relative to VQ CPC, demonstrating that learning the discrete representations results in better recognition.
		The best performing technique for each dataset is denoted in \textbf{\darkgreen{bold}}.
	}
	\begin{tabular}{ccccccc}
		\toprule
		& \multicolumn{2}{c}{Wrist} & \multicolumn{2}{c}{Waist} & \multicolumn{2}{c}{Leg} \\ 
		\multirow{-2}{*}{Method} & HHAR & Myogym & Mobiact & Motionsense & MHEALTH & PAMAP2 \\ 
		\midrule
		SAX + LSTM class. & 43.78 $\pm$ 0.96 & 14.22 $\pm$ 0.40 & 66.45 $\pm$ 0.18 & 70.14 $\pm$ 0.36 & 41.04 $\pm$ 0.58 & 47.61 $\pm$ 1.04 \\ 
		SAX-REPEAT + LSTM class. & 39.17 $\pm$ 0.69 & 28.65 $\pm$ 0.69 & 71.73 $\pm$ 0.74 & 71.31 $\pm$ 0.74 & 38.89 $\pm$ 0.48 & 43.30 $\pm$ 1.83 \\ 
		\hdashline\noalign{\vskip 0.5ex}
		VQ CPC + LSTM class.  & \textbf{\darkgreen{60.76 $\pm$ 1.09}} & 29.62 $\pm$ 0.52 & 76.34 $\pm$ 0.30 & 89.06 $\pm$ 0.24 & 48.86 $\pm$ 0.34 & 55.28 $\pm$ 0.34 \\ 
		VQ CPC + GRU class.  & 60.26 $\pm$ 0.83 & \textbf{\darkgreen{31.65 $\pm$ 0.29}} & \textbf{\darkgreen{77.78 $\pm$ 0.17}} & \textbf{\darkgreen{89.23 $\pm$ 0.23}} & \textbf{\darkgreen{49.01 $\pm$ 0.30}} & \textbf{\darkgreen{56.92 $\pm$ 0.26}} \\ 
		\bottomrule
	\end{tabular}
	\label{tab:other_discretization_methods}
\end{table*}

The activity recognition capabilities of discrete representation learning are shown in Tab.\ \ref{tab:diff_locs}, obtaining the highest performance on three benchmark datasets.
In this experiment, we compare the performance to SAX, which is an established method for discretizing uni-variate time-series data, and SAX-REPEAT \cite{mohammad2014robust}, which utilizes SAX for discretizing multi-channel time-series data.
For appropriate comparison, SAX also results in one symbol for every second timestep of sensor data, with an alphabet size of $512$. 
SAX-REPEAT separately applies SAX to each channel of accelerometer data, resulting in tuples of indices for every second timestep. 
As utilizing the tuples as-is results in a possible dictionary size of $512^3$, SAX-REPEAT performs K-Means clustering (with k=512) on the tuples in order to maintain an alphabet size of $512$, where the cluster indices function as the discrete representation.
The same classifier setup (Sec.\ \ref{sec:classifier_network}) is utilized for activity recognition (including the parameter tuning for classification) and five random runs of the five fold validation F1-score is detailed in Tab.\ \ref{tab:other_discretization_methods}.
The comparison is drawn against the learned discrete representation method, which is VQ-CPC.

For all datasets, the SAX baseline performs poorly compared to the learned discrete representations, showing a reduction of over 10\% for HHAR, Myogym, and Motionsense, and a smaller reduction for Mobiact, MHEALTH, and PAMAP2.
This can be expected as SAX utilizes the magnitude of the accelerometer data as the input, thereby reducing three channels to one and losing information about direction of movement.
Considering SAX-REPEAT next, we see that it shows worsened performance to SAX on HHAR, MHEALTH, and PAMAP2.
For Mobiact, the performance is only 6\% lower than VQ-CPC + GRU classifier, whereas for the other datasets, the difference is greater.
Only on Myogym, the performance is better than VQ-CPC, albeit substantially lower than the state-of-the-art self-supervised as well as end-to-end training methods. 
The lower performance for SAX and SAX-REPEAT for Myogym also indicates that discretization is not a good option for fine-grained activities.
\edit{
	Our experiments clearly show that SAX and SAX-REPEAT are worse at recognizing activities compared to VQ-CPC.
	Further, the reduction in performance of SAX-REPEAT relative to SAX on HHAR, MHEALTH, and PAMAP2, indicates that modifying SAX to apply to multi-variate data is challenging.
}
Overall, Tab.\ \ref{tab:other_discretization_methods} shows that the traditional methods are not effective for discretizing acclerometer data, and that learning a codebook in an unsupervised, data-driven way results in a better mapping of sensor data to discrete representations.

\subsection{Effect of the Learned Alphabet Size}
\begin{table*}[!t]
	\centering
	\small
	\caption{
		Studying the impact of the maximum dictionary size on activity recognition:
		we explicitly limit the size to $\{32, 64, 128, 256, 512\}$ codebook vectors and study how performance is affected by the applied constraint.
		The mean dictionary size across the five folds for the base setup of VQ CPC + GRU classifier is shown in brackets in the last row.
		For HHAR, we observe a substantial increase of over 7\% by limiting the size to 64.
		For MHEALTH and PAMAP2, the improvements are more modest.
		This indicates that more deliberate choice of dictionary size can result in further performance increases.
	}
	\begin{tabular}{P{0.2\textwidth} P{0.11\textwidth} P{0.11\textwidth} P{0.11\textwidth} P{0.11\textwidth} P{0.11\textwidth} P{0.11\textwidth}}
		\toprule
		& \multicolumn{2}{c}{Wrist} & \multicolumn{2}{c}{Waist} & \multicolumn{2}{c}{Leg} \\ 
		\multirow{-2}{*}{Max. dict. size} & HHAR & Myogym & Mobiact & Motionsense & MHEALTH & PAMAP2 \\ 
		\midrule
		32 & 11.58 $\pm$ 0.12 & 2.80 $\pm$ 0.00 & 29.89 $\pm$ 0.24 & 37.46 $\pm$ 0.26 & 14.38 $\pm$ 0.28 & 21.26 $\pm$ 0.37 \\ 
		64 & \textbf{\darkgreen{67.62 $\pm$ 0.21}} & 20.78 $\pm$ 0.28 & 74.19 $\pm$ 0.19 & 89.70 $\pm$ 0.18 & 48.77 $\pm$ 0.36 & \textbf{\darkgreen{58.06 $\pm$ 0.51}} \\ 
		128 & 57.71 $\pm$ 0.87 & 27.81 $\pm$ 0.28 & 75.73 $\pm$ 0.41 & 79.49 $\pm$ 0.24 & \textbf{\darkgreen{49.62 $\pm$ 0.51}} & 56.56 $\pm$ 0.66 \\ 
		256 & 60.21 $\pm$ 0.66 & 18.13 $\pm$ 0.33 & 75.53 $\pm$ 0.24 & \textbf{\darkgreen{90.28 $\pm$ 0.28}} & 47.71 $\pm$ 0.49 & 57.12 $\pm$ 0.34 \\ 
		512 & 60.41 $\pm$ 0.55 & 12.92 $\pm$ 0.44 & 64.25 $\pm$ 0.51 & 71.73 $\pm$ 0.23 & 46.81 $\pm$ 0.80 & 53.40 $\pm$ 0.60 \\  	
		\hdashline\noalign{\vskip 0.5ex}
		VQ CPC + GRU classifier  & 60.26 $\pm$ 0.83 (127.8) & \textbf{\darkgreen{31.65 $\pm$ 0.29}} (155) & \textbf{\darkgreen{77.78 $\pm$ 0.17}} (148) & 89.23 $\pm$ 0.23 (140) & 49.01 $\pm$ 0.30 (130.2) & 56.92 $\pm$ 0.26 (140.4) \\  
		\bottomrule
	\end{tabular}
	\label{tab:diff_dictionary_sizes}
\end{table*}

One of the advantages of discrete representation learning via Vector Quantization is the control over the size of the learned dictionary.
It can be set depending on the required fidelity of the learned representations and the capacity of the computation power available for classification.
For applications where the separation of activities requires a small dictionary (e.g., 8 or 16 symbols), we can accordingly set the dictionary size and thereby save computation power during classification.
For our base setup (Tab.\  \ref{tab:diff_locs}), we utilize indepedent quantization of partitions of the vectors, resulting in a possible $100^2$ dictionary size. 
Here, we explicitly control the dictionary size by setting number of groups = 1 and vary the number of variables (i.e., number of codebook vectors) between (32, 64, 128, 256, 512).
We also note that the final dictionary size can be lower than the codebook size and depends on the underlying movements and sensor data.
We perform activity recognition on the resulting discrete representations of windows of sensor data using the best performing models from Tab.\ \ref{tab:diff_locs}, albeit with increasing alphabet sizes.
The results from this experiment are tabulated in Tab.\ \ref{tab:diff_dictionary_sizes}.
A similar analysis was also performed in VQ-APC \cite{chung2020vector}.

First, we notice that having a max.\ alphabet size of 32 results in poor performance. 
Such a small dictionary size provides limited descriptive power for the representations and therefore leads to significant drops in performance relative to the base setup of utilizing multiple groups during quantization (see Sec.\ \ref{sec:prev_mode_collapse}).
Along the same lines, having too large a dictionary size is also slightly detrimental (max dict size = 512), as it can lead to long-tailed distributions of the symbolic representations and the network starting to pay attention to noises instead.

We obtain the highest performance when the max dictionary sizes are 64, 128, or 256.
For HHAR, the constraint on the dictionary size results in an increase of over 7\% relative to the base setup (VQ CPC + GRU classifier).
For Myogym and Mobiact however, not constraining the resulting dictionary sizes is the best option, with clear increases over the constrained models. 
For Motionsense and PAMAP2, controlling the learned alphabet size results in modest performance improvements of 1\%, whereas for MHEALTH, it is around 0.6\%. 
Clearly, with reducing codebook sizes, the model is forced to choose what information to discard and what to encode \cite{chung2020vector}.
This process can result in higher performance as the network can more efficiently learn to ignore irrelevant information (such as noise) and picks up more discriminatory information. 

Next, we consider the mean dictionary size across all folds obtained by utilizing groups = 2 (as in Tab. \ref{tab:diff_locs}, see Sec. \ref{sec:prev_mode_collapse} for reference).
For all target datasets, the size < 160 symbols, emphasizing that effective recognition can be obtained using just around 130-160 symbols. 
This is encouraging, as downstream tasks such as gesture or activity spotting, can be performed more easily with a smaller dictionary size. 
The importance of creating groups during discretizarion is also visible, as it results in the highest performance for two target datasets, along with comparable performance for three datasets, without having to further tune the dictionary size as a hyperparameter.

\subsection{\edit{Impact of the Encoder's Output Frequency}}
\label{sec:symbol_duration}
\begin{table*}[!t]
	\centering
	\small
	\caption{
	Investigating the impact of the encoder's output frequency: the base setup from Sec. \ref{sec:method} results in a discrete symbol for approx.\ every second timestep of sensor data. 
	\edit{
		By adjusting the encoder architecture, we study whether an output frequency of 50 Hz (same as input) or 11.5 Hz (approx.\ halved relative to the base setup) is more appropriate for the representations.
		We observe that the base setup of 24.5 Hz results in better performance while also reducing computational costs.		
	}
	}
	\begin{tabular}{ccccccc}
		\toprule
		& \multicolumn{2}{c}{Wrist} & \multicolumn{2}{c}{Waist} & \multicolumn{2}{c}{Leg} \\ 
		%
		\multirow{-2}{*}{Encoder output freq.} & HHAR & Myogym & Mobiact & Motionsense & MHEALTH & PAMAP2 \\ 
		\midrule
		50 Hz & 58.02 $\pm$ 0.46 & 18.30 $\pm$ 0.45 & 77.65 $\pm$ 0.43 & 85.25 $\pm$ 0.39 & 48.43 $\pm$ 0.66 & 51.36 $\pm$ 0.91 \\ 
		11.5 Hz & 48.95 $\pm$ 1.26 & 16.50 $\pm$ 0.31 & 53.51 $\pm$ 0.26 & 63.68 $\pm$ 0.37 & 32.78 $\pm$ 0.66 & 41.62 $\pm$ 0.61 \\  	
		\hdashline\noalign{\vskip 0.5ex}
		24.5 Hz  & \textbf{\darkgreen{60.26}} $\pm$ 0.83 & \textbf{\darkgreen{31.65 $\pm$ 0.29}} & \textbf{\darkgreen{77.78 $\pm$ 0.17}} & \textbf{\darkgreen{89.23}} $\pm$ 0.23 & \textbf{\darkgreen{49.01 $\pm$ 0.30}} & \textbf{\darkgreen{56.92 $\pm$ 0.26}} \\ 
		\bottomrule
	\end{tabular}
	\label{tab:input_downsampling}
\end{table*}

In Sec.\ \ref{sec:method}, we detailed the architecture for learning discrete representations of human movements.
The convolutional encoder results in \edit{approximately} one latent representation per two timesteps of sensor data.
\edit{
	With appropriate architectural modifications, we can increase or reduce the output frequency of the encoder.
	Intuitively, a lower output frequency can be problematic as too much motion (and variations of motion) can get mapped to each symbol.
}
When this occurs, nuances in movements are not captured well by the symbolic representations.
\edit{
	In this experiment, we vary the output frequency and study the impact on performance. 
}
The convolutional encoder is modified accordingly:
\edit{
	\emph{i)} for an output frequency of 50 Hz (i.e., no downsampling relative to the input), we change the stride of the first block to 1 and for the second block, set the kernel size and stride = 1; and
	\emph{ii)} for an output frequency of 11.5 Hz (i.e., further downsampling by two relative to the base setup), the second block also has a kernel size of 4 with stride = 2. 
}
We perform activity recognition on the six target datasets, and report the five-fold cross validation performance across five randomized classification runs in Tab.\ \ref{tab:input_downsampling}.

\edit{
	As expected, an encoder output frequency of 11.5 Hz (i.e., \# timesteps / symbol $\approx$ 4) results in substantial reductions in performance relative to the base setup (where the output frequency is 24.5 Hz).
}
For HHAR, the drop in performance is around 10\%. 
However, for Myogym, MHEALTH, and PAMAP2, it is over 15\%. 
The waist-based datasets see the highest impact on performance, experiencing a reduction of over 20\% with the longer duration mapping.
We can reasonably expect that \edit{a lower encoder output frequency}, will result in further reduction in performance.

We also note that \edit{maintaining the same output frequency as the input} also causes a drop, albeit smaller, in the test set performance.
While this configuration can be utilized for obtaining discrete representations, the training times are considerably higher, while also not resulting in performance improvements. 
Therefore, \edit{an output frequency of 24.5 Hz (relative to an input of 50 Hz)} is better, allowing for quicker training while also covering more of the underlying motion.

\subsection{NLP-based pre-training with RoBERTa}

\begin{figure}[!t]
	\centering
	\includegraphics[height=3.0cm]{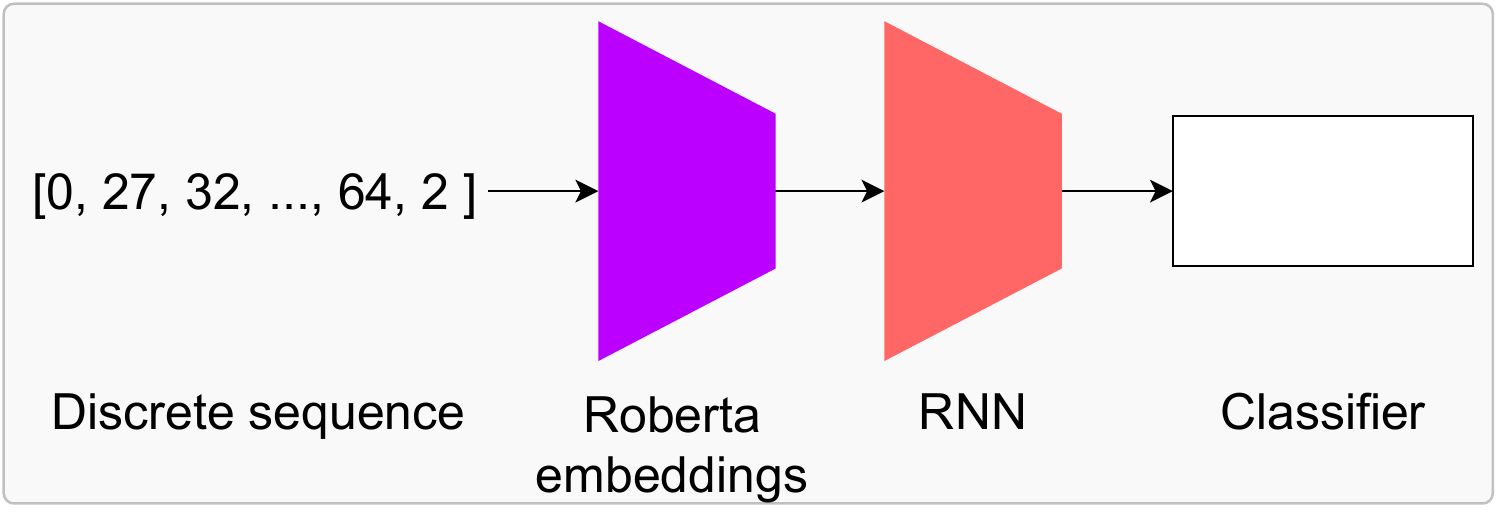}
	\caption{
		\edit{
		Using RoBERTa embeddings for classifying discrete representations: first, we pre-train a RoBERTa model on discrete representations (indexed) from Capture-24.
		Subsequently, we replace the learnable embeddings with frozen RoBERTa embeddings in order to study the impact of the NLP-based pre-training.
		A GRU network is used along with an MLP for activity recognition.
		}
	}
	\label{fig:classification_with_roberta}
\end{figure}

\begin{table*}[!t]
	\centering
	\small
	\caption{
		Evaluating the impact of utilizing pre-trained RoBERTa embeddings (obtained from the discretized Capture-24 dataset) for recognizing activities: we compare the performance to random learnable embeddings (VQ CPC) as well as other self-supervised and end-to-end baselines.
		All scores apart from RoBERTa small/medium are taken from Tab. \ref{tab:diff_locs}.
		The best performing technique overall for each dataset is denoted in \textbf{\darkgreen{bold}}, whereas the best unsupervised method is shown using $^\dagger$.
		Therefore, methods with \textbf{\darkgreen{bold}}$^\dagger$ are the best method overall and the best unsupervised method as well.
		The performance across five random runs for five fold test set F1-scores is detailed below.
		We observe that the addition of RoBERTa embeddings results in improvements for \textbf{all} target datasets.
		Further, we achieve the state-of-the-art for three datasets, which cover locomotion and daily living activities.
	}
	\begin{tabular}{P{0.15\textwidth}cccccc}
		\toprule
		& \multicolumn{2}{c}{Wrist} & \multicolumn{2}{c}{Waist} & \multicolumn{2}{c}{Leg} \\ 
		\multirow{-2}{*}{Method} & HHAR & Myogym & Mobiact & Motionsense & MHEALTH & PAMAP2 \\ 
		\midrule
		\rowcollight \multicolumn{7}{c}{Supervised baselines} \\ 
		Conv. classifier$^*$ & 55.63 $\pm$ 2.05 & 38.21 $\pm$ 0.62 & 78.99 $\pm$ 0.38 & 89.01 $\pm$ 0.89 & 48.71 $\pm$ 2.11 & 59.43 $\pm$ 1.56 \\ 
		DeepConvLSTM$^*$ & 52.37 $\pm$ 2.69 & 39.36 $\pm$ 1.56 & \textbf{\darkgreen{82.36 $\pm$ 0.42}} & 84.44 $\pm$ 0.44 & 44.43 $\pm$ 0.95 & 48.53 $\pm$ 0.98 \\ 
		GRU classifier$^*$ & 45.23 $\pm$ 1.52 & 36.38 $\pm$ 0.60 & 75.74 $\pm$ 0.60 & 87.42 $\pm$ 0.52 & 44.78 $\pm$ 0.47 & 54.35 $\pm$ 1.64 \\
		\rowcollight \multicolumn{7}{c}{Self-supervision + MLP classifier} \\ 
		M-task self.\ sup$^*$ & 57.55 $\pm$ 0.75 & 42.73 $\pm$ 0.49 & 72.17 $\pm$ 0.38 & 86.15 $\pm$ 0.42 & 50.39 $\pm$ 0.72 & \textbf{\darkgreen{60.25 $\pm$ 0.72}}$^\dagger$ \\
		SimCLR$^*$ & 56.34 $\pm$ 1.28 & \textbf{\darkgreen{47.82 $\pm$ 1.03}}$^\dagger$ & 75.78 $\pm$ 0.37 & 87.93 $\pm$ 0.61 & 42.11 $\pm$ 0.28 & 58.38 $\pm$ 0.44 \\ 
		Enhanced CPC$^*$ & 59.25 $\pm$ 1.31 & 40.87 $\pm$ 0.50 & 78.07 $\pm$ 0.27 & 89.35 $\pm$ 0.32 & \textbf{\darkgreen{53.79 $\pm$ 0.83}}$^\dagger$ & 58.19 $\pm$ 1.22 \\ 
		%
		\rowcollight \multicolumn{7}{c}{Discrete representations + GRU classifier} \\ 
		VQ CPC   & 60.26 $\pm$ 0.83 & 31.65 $\pm$ 0.29 & 77.78 $\pm$ 0.17 & 89.23 $\pm$ 0.23 & 49.01 $\pm$ 0.30 & 56.92 $\pm$ 0.26 \\ 
		VQ CPC + RoBERTa small & 62.30 $\pm$ 0.68 & 34.41 $\pm$ 0.45 & 79.42 $\pm$ 0.38$^\dagger$ & \textbf{\darkgreen{91.76 $\pm$ 0.22}}$^\dagger$ & 51.77 $\pm$ 0.28 & 59.34 $\pm$ 0.48 \\ 
		
		VQ CPC + RoBERTa medium & \textbf{\darkgreen{63.31 $\pm$ 0.38}}$^\dagger$ & 35.16 $\pm$ 0.29 & 78.99 $\pm$ 0.33 & 91.45 $\pm$ 0.26 & 51.59 $\pm$ 0.42 & 59.53 $\pm$ 0.38 \\ 
		\bottomrule
	\end{tabular}
	
	\label{tab:adding_roberta}
\end{table*}

One of the advantages of converting the sensor data into discrete sequences, is that it allows us to apply powerful NLP-based pre-training techniques such as BERT \cite{devlin2018bert}, RoBERTa \cite{liu2019roberta}, GPT \cite{radford2018improving}, etc., as learned embeddings for the RNN classifier. 
In addition, the release of new techniques for text-based self-supervision can be accompanied by corresponding updates to the classification on the discrete representations learned from movement data. 
Therefore, in this experiment, we investigate whether Robustly Optimized BERT Pretraining Approach (RoBERTa) \cite{liu2019roberta} based pre-training on the symbolic representations is useful for improving activity recognition performance.
\edit{
	While RoBERTa can increase the computational footprint of the recognition system, it can be potentially replaced with recent advancements in distilling and pruning BERT models such as SNIP \cite{lin2020pruning}, ALBERT \cite{lan2019albert}, and DistillBERT \cite{sanh2019distilbert} while maintaining similar performance.
}

First, we extract the symbolic representations on the large-scale Capture-24 dataset (utilizing 100\% of the train split), and use it to pre-train two RoBERTa models, called `small' and `medium'.
The `small' model contains an embedding size of $128$ units, a feedforward size of 512 units, and 2 Transformer \cite{vaswani2017attention} encoder layers with 8 heads each.
On the other hand, the `medium' sized model comprises embeddings of size $256$, with a feedforward dimension of 1,024, and 4 Transformer encoder layers with 8 heads each.
The aim of training models with two different sizes is to investigate whether increased depth results in corresponding performance improvements or not.
Following the protocol from Tab. \ref{tab:diff_locs}, the performance across five random runs is reported for five-fold cross validation is reported in Tab. \ref{tab:adding_roberta}.
\edit{
	As shown in Fig.\ \ref{fig:classification_with_roberta}, the randomly initialized learnable embedding layer is replaced with the learned RoBERTa models, which are frozen.
	Only the GRU classifier is updated with label information during the classifier training.
}

First, we observe that utilizing the learned RoBERTa embeddings (VQ CPC + RoBERTa small/medium in Tab. \ref{tab:adding_roberta}) instead of the random learnable embeddings (VQ CPC in Tab. \ref{tab:adding_roberta}) results in performance improvements for \textbf{all} target datasets.
This indicates the positive impact of pre-training with RoBERTa. 
For the small version, the wrist-based HHAR and Myogym see increases of 2\% and 2.8\% respectivelty.
A similar trend is observed for the waist-based datasets as well, improving by 1.6\% and 2.5\% for Mobiact and Motionsense. 
Finally, the leg-based datasets also see improvements of around 2.5\% each. 
Interestingly, the medium sized model of RoBERTa shows a similar performance to the small version, except for the wrist-based HHAR and Myogym, where the increase over random embeddings is 3\% and 3.5\% respectively. 
The similar performance demonstrated by the medium version indicates that the increase in model size did not result in corresponding performance improvements, likely because Capture-24 is not large enough to leverage the bigger architecture. 
Potentially, an even larger dataset (e.g., Biobank \cite{doherty2017large}) can be utilized for the medium version (or even larger variants). 

The advantage of performing an additional round of pre-training via RoBERTa is clearly observed in Tab. \ref{tab:adding_roberta}, as VQ CPC + RoBERTa outperforms the state-of-the-art for representation learning on three datasets (HHAR, Mobiact, and Motionsense) by clear margins.
For the leg-based datasets, the performance with the addition of RoBERTa is closer to the most effective methods, through improved learning of embeddings.
This result is promising for wearables applications, as it proves that the rapid advancements from natural language processing can be applied for improved activity recogntion as well.

\section{Discussion}
\label{sec:discussion}
In this paper, we propose a return to discrete representations as descriptors of human movements for wearables-based applications.
Going beyond prior works such as SAX, we instead learn the mapping between short spans of sensor data and the symbolic representations.
\edit{In what follows, we will first} visualize the distributions of the discrete representations for activities across the target datasets, and examine the similarities and differences.
The latter half of this section contains an introspection of the method itself, along with the lessons learned during our explorations.

\subsection{Visualizing the Distributions of the Discrete Representations}
\begin{figure} 
	\centering
	\includegraphics[width=0.8\textwidth]{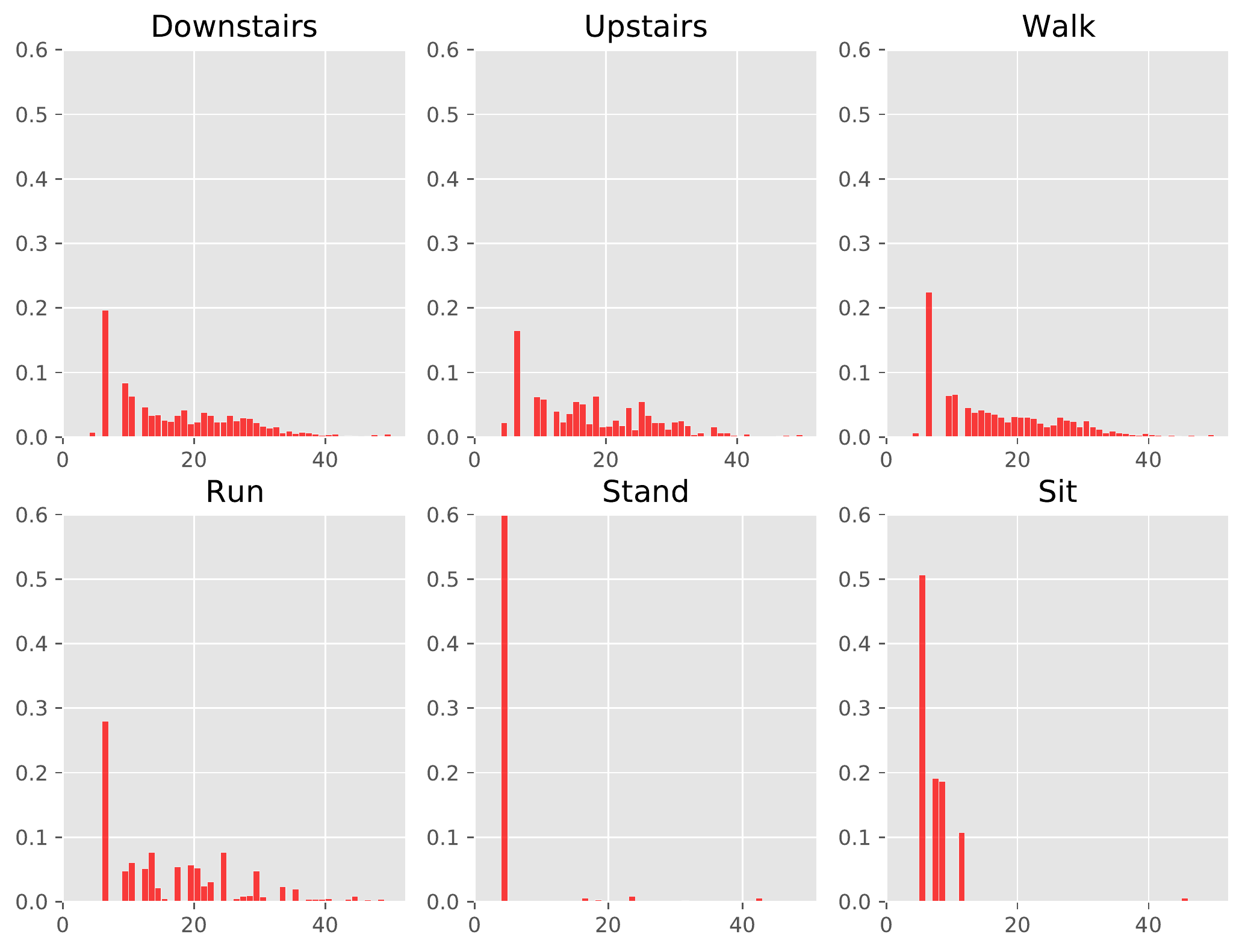}
	\caption{
	Plotting the histograms per class of the derived discrete representations for the train set from the first fold of Motionsense.
	The y-axis corresponds to the fraction of the number of occurrences for the symbols computed against the available data.
	The x-axis comprises the discrete representations, which are numbered.
	We observe that standing and sitting are covered by only a few symbols, which can be reasoned by the lack of movements in these classes.
	Walking up and down the stairs are also similar, yet hard to distinguish via visual examination from both walking and running.
	For clarity, we truncate the y-axis to 0.6 and x-axis to 50 symbols, in order to show the plot in more detail.
	The full figure is available in the Appendix (refer to Fig. \ref{fig:discrete_rep_hist_big}).
	}
	\label{fig:discrete_rep_hist}
\end{figure}

Through our experiments (results reported in Tab. \ref{tab:diff_locs} and \ref{tab:adding_roberta}), we quantitatively established the performance of the learned discrete representations.
We demonstrated that training GRU classifiers with randomly initialized embeddings (Tab.\ \ref{tab:diff_locs}) results in effective activity recognition on five of six benchmark datasets.
In addition, deriving pre-training embeddings from the discrete representations via RoBERTa further pushes the performance, exceeding state-of-the-art self-supervision on three datasets.
Given their recognition capabilities, we plot the distributions of the discrete representations for each activity, in order to visualize how the underlying movements may be different.
\edit{
This serves as a first check to visually examine whether similar activities such as walking and walking up/ downstairs--which may have similar underlying movements--are actually represented in similar discrete representations. 
}

\begin{figure}[!h]
	\centering
	\includegraphics[width=\textwidth]{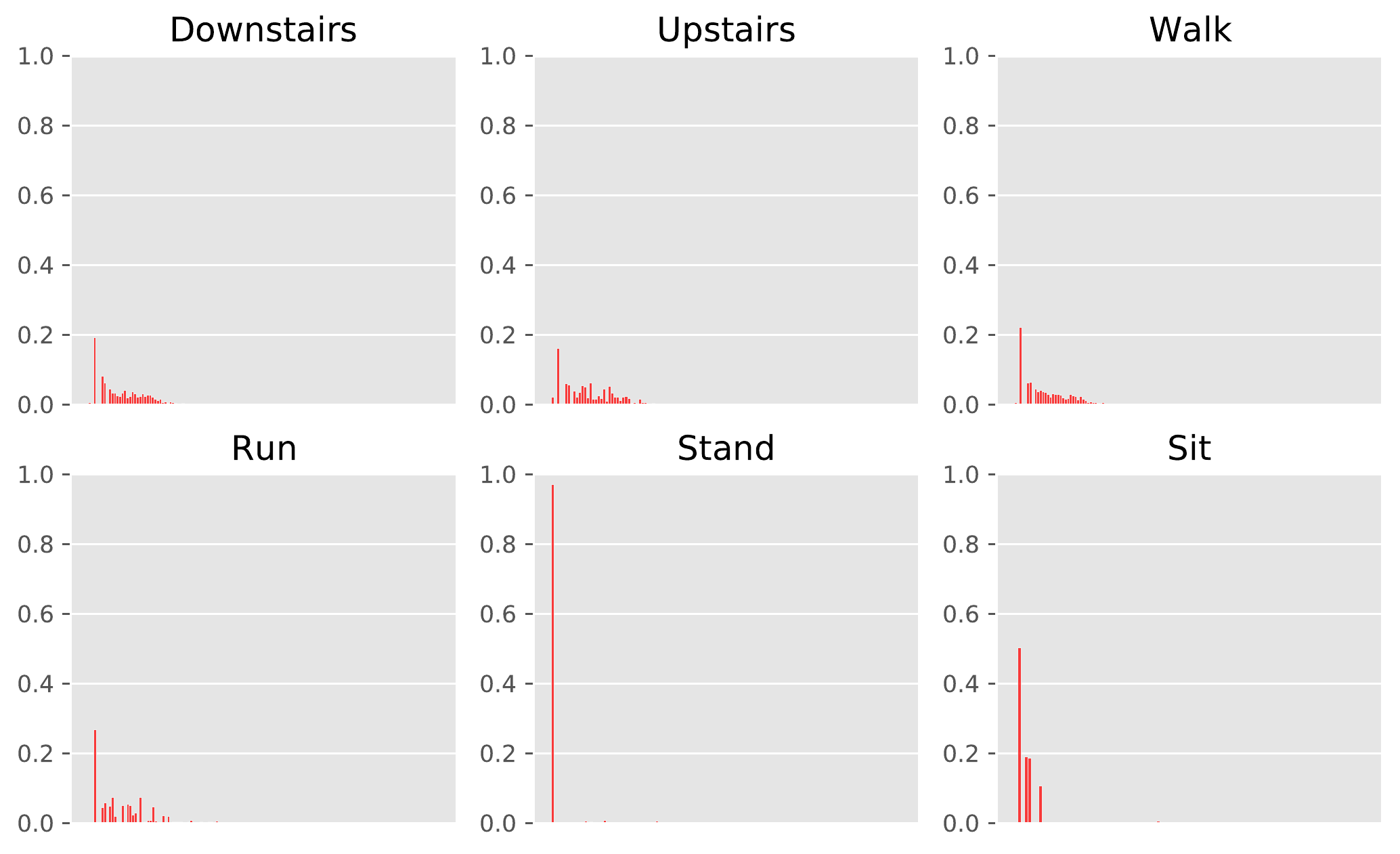}
	\caption{
		Visualizing the histograms of discrete representations per class for the first fold of Motionsense. 
		This is the full figure from which Fig. \ref{fig:discrete_rep_hist} is obtained by truncating the x-axis to 50 symbols. 	
	}
	\label{fig:discrete_rep_hist_big}
\end{figure}

In Fig. \ref{fig:discrete_rep_hist} and \ref{fig:discrete_rep_hist_big}, we present the histograms of the discrete representations per activity. 
The y-axis comprises the fraction of the total sum of representations held by each discrete symbol.
First, we note that the discrete representations exhibit long-tailed distributions, with a significant portion of representations being used very sparsely (more clearly visible in Fig.\ \ref{fig:discrete_rep_hist_big}). 
The impact of such a distribution is challenging to predict -- on one hand, the rarely occuring symbols can increase the complexity of the classifiers (and embeddings) due to their numerosity, while on the other, it is likely they capture more niche movements as they may be performed by participants. 
Such niche movements can potentially help with classification of less frequent activities.
In addition, we also note that the distributions for sitting and standing contain limited variability, as the underlying movements themselves exhibit less motion. 
This somewhat verifies that the learned discrete representations correspond to the movements themselves, given that a lack of movement is captured in the distribution of representations per activity.
Interestingly, the histograms for going up and down the stairs look very similar, while walking also retains similarities to them. 
Running looks slightly different, with more spreading out across the symbolic representations, therefore indicating that higher variability of underlying movements involving running, which makes sense intuitively.
Therefore, the distributions of the discrete representations provide practitioners with an additional tool for understanding human activities as well as the underlying movements. 
This is a point in favor of discrete representations, as the analysis is possible in conjunction with comparable if not better performance for activity recognition.

\subsection{\edit{Analyzing the VQ-CPC Discretization Framework}}
\edit{
	In the previous sections, we established the effectiveness of our VQ-CPC-based framework for learning discrete representations of human movements. 
	Here, we examine specific components of the framework, in order to understand their impact on both discrete representation learning, as well as on downstream activity recognition.
	To this end, we consider the following components:
	\emph{(i) the Encoder network -- } where the architecture determines the duration of time covered by each symbol; 
	and
	\emph{(ii) the self-supervised pretext task --} which acts as the base for the discrete representation learning and is used for learning the quantization codebook. 
	In what follows, we study the activity recognition performance for the same target datasets, albeit replacing the aforementioned components of the framework with suitable alternatives and examine the performance.
}

\subsubsection{\edit{Impact of the Encoder Architecture}}
\label{sec:impact_of_encoder_arch}
\begin{table*}[!t]
	\centering
	\small
	\caption{
		\edit{
		Studying the impact of the encoder network architecture on recognition performance: we examine whether a smaller receptive field in the encoder network is more preferable for discrete representation learning.
		We note that larger receptive fields (e.g., 8 or 16) result in performance reduction compared to the base setup (filter size = 4). 
		}
	}
	\begin{tabular}{P{0.24\textwidth}cccccc}
		\toprule
		& \multicolumn{2}{c}{Wrist} & \multicolumn{2}{c}{Waist} & \multicolumn{2}{c}{Leg} \\ 
		%
		\multirow{-2}{*}{Encoder arch.} & HHAR & Myogym & Mobiact & Motionsense & MHEALTH & PAMAP2 \\ 
		\midrule
		Base setup (i.e., filt. size = 4) & 60.26 $\pm$ 0.83 & 31.65 $\pm$ 0.29 & 77.78 $\pm$ 0.17 & 89.23 $\pm$ 0.23 & 49.01 $\pm$ 0.30 & 56.92 $\pm$ 0.26 \\ 
		\hdashline\noalign{\vskip 0.5ex}
		Base setup + filt. size = 8 & 55.15 $\pm$ 0.60 & 16.80 $\pm$ 0.44 & 70.15 $\pm$ 0.39 & 77.95 $\pm$ 0.28 & 47.73 $\pm$ 0.26 & 43.68 $\pm$ 0.52 \\  		
		Base setup + filt. size = 16 & 49.07 $\pm$ 0.45 & 16.59 $\pm$ 0.42 & 71.22 $\pm$ 0.12 & 81.38 $\pm$ 0.56 & 46.18 $\pm$ 0.37 & 44.29 $\pm$ 0.50 \\  	
		\hdashline\noalign{\vskip 0.5ex}
		Multi-task enc. \cite{saeed2019multi} & 45.32 $\pm$ 1.43 & 22.31 $\pm$ 0.11 & 48.92 $\pm$ 0.23 & 76.37 $\pm$ 0.41 & 39.60 $\pm$ 0.76 & 47.93 $\pm$ 0.76 \\  	
		\bottomrule
	\end{tabular}
	\label{tab:diff_encoder_arch}
\end{table*}

\edit{
	Our Encoder architecture is based on the Enhanced CPC framework \cite{haresamudram2023investigating}.
	As detailed in Sec.\ \ref{sec:disrete_rep_learning_setup}, it contains four convolutional blocks, with a kernel size of (4,1,1,1), and stride of (2,1,1,1), respectively.
	Therefore, the resulting $z$-vectors are obtained approximately once every second timestep (we obtain 49 $z$-vectors from an input window of 100 timesteps due to the striding).
	From Tab.\ \ref{tab:input_downsampling}, we see that decreasing the encoder output frequency to 11.5 Hz is detrimental to performance. 
	We now conduct a deeper analysis into the design of suitable encoders by considering the following configurations:
	\emph{(i)} increasing the kernel size of the first layer to (8, 16), while keeping the architecture otherwise identical to the base setup; and
	\emph{(ii)} utilizing an encoder identical to the convolutional encoders of Multi-task self-supervision \cite{saeed2019multi} and SimCLR \cite{tang2020exploring}.
	The learning rate and L2 regularization are identical to the base setup, and we also tune the number of aggregator layers across (2,4,6) layers, as described in Sec.\ \ref{sec:settings}.
	The results from this analysis are tabulated in Tab.\ \ref{tab:diff_encoder_arch}.
}

\edit{
	In the base setup, movement across four timesteps (0.08s at 50 Hz) contributes to each symbol.
	This is increased to (8, 16) timesteps depending on the filter size of the first layer.
	From Tab.\ \ref{tab:diff_encoder_arch}, we see that this is detrimental to activity recognition.
	Clearly, it becomes difficult to learn the mapping between 8 timesteps (and greater, i.e., $\geq 0.16s$) to discrete representations, as the underlying movements are covering much longer durations and thus become too coarse for symbolic representations.
	We extend this analysis further by applying the encoder from Multi-task self-sup. \cite{saeed2019multi} instead of the base encoder for learning discrete representations.
	As the encoder contains three blocks with filter sizes of (24, 16, 8), a total of 46 time steps (i.e., 0.92 seconds of movements) contribute to each symbol.
	We observe a significant drop in performance as a result, with around 15\% reduction for HHAR and approx.\ 30\% decrease for Mobiact.
	While it is preferable to learn symbols that represent short spans of time, clearly, accurately mapping longer durations to symbols is a difficult proposition.
	From our exploration, a filter size of 4 (for the first layer) seems ideal, covering sufficient motion as well as resulting in accurate activity recognition.
	This also motivates the architecture of our encoder, where all layers apart from the first one have a filter size of 1. 
	Having multiple layers (after the first) with filter size $>1$ would result in $z$-vectors corresponding to longer durations, thereby resulting in reduced performance.
}

\subsubsection{\edit{Effect of the Base Self-Supervised Method on Recognition Performance}}

\begin{table*}[!t]
	\centering
	\small
	\caption{
		\edit{
		Evaluating the impact of the base self-supervised method on activity recognition: we study the effect of utilizing other self-supervised methods -- Autoencoders, Multi-task self-sup.\ learning, and SimCLR -- in lieu of Enhanced CPC for model training.
		We observe that Enhanced CPC (i.e., which is the base self-supervision for VQ-CPC) is clearly the most suitable self-supervised method, while simpler techniques such as Autoencoders can also be utilized, albeit with performance reductions. 
		Further, we also investigate whether the VQ-CPC encoder can result in better performance for other methods, and find that to be true for both Multi-task self-supervision and SimCLR.
		}
	}
	\begin{tabular}{P{0.22\textwidth}cccccc}
		\toprule
		& \multicolumn{2}{c}{Wrist} & \multicolumn{2}{c}{Waist} & \multicolumn{2}{c}{Leg} \\ 
		\multirow{-2}{*}{Base self-supervision} & HHAR & Myogym & Mobiact & Motionsense & MHEALTH & PAMAP2 \\ 
		\midrule
		VQ-CPC & 60.26 $\pm$ 0.83 & 31.65 $\pm$ 0.29 & 77.78 $\pm$ 0.17 & 89.23 $\pm$ 0.23 & 49.01 $\pm$ 0.30 & 56.92 $\pm$ 0.26 \\ 
		\hdashline\noalign{\vskip 0.5ex}
		VQ-Autoencoder & 48.12 $\pm$ 0.72 & 30.61 $\pm$ 0.97 & 73.03 $\pm$ 0.59 & 79.59 $\pm$ 0.68 & 45.61 $\pm$ 0.81 & 48.99 $\pm$ 1.37 \\  	
		VQ-Autoencoder + VQ-CPC encoder & 50.60 $\pm$ 1.05 & 33.15 $\pm$ 0.44 & 73.26 $\pm$ 0.47 & 75.89 $\pm$ 0.62 & 37.75 $\pm$ 0.59 & 51.23 $\pm$ 0.66 \\  	  	
		\hdashline\noalign{\vskip 0.5ex}
		VQ-Multi-task self-sup. & 45.32 $\pm$ 1.43 & 22.31 $\pm$ 0.11 & 48.92 $\pm$ 0.23 & 76.37 $\pm$ 0.41 & 39.60 $\pm$ 0.76 & 47.93 $\pm$ 0.76 \\  	
		VQ-Multi-task self-sup. + VQ-CPC encoder & 58.29 $\pm$ 0.69 & 24.53 $\pm$ 0.62 & 72.87 $\pm$ 0.14 & 80.49 $\pm$ 0.60 & 44.85 $\pm$ 0.63 & 56.28 $\pm$ 0.40 \\
		\hdashline\noalign{\vskip 0.5ex}
		VQ-SimCLR & 25.34 $\pm$ 0.40 & 11.17 $\pm$ 0.17 & 14.24 $\pm$ 0.38 & 60.67 $\pm$ 0.12 & 9.04 $\pm$ 0.09 & 19.62 $\pm$ 0.66 \\  	
		VQ-SimCLR + VQ-CPC encoder & 58.49 $\pm$ 0.29 & 2.85 $\pm$ 0.01 & 59.62 $\pm$ 0.35 & 59.54 $\pm$ 0.15 & 41.62 $\pm$ 0.32 & 55.80 $\pm$ 0.52 \\
		\bottomrule
	\end{tabular}
	\label{tab:diff_self_sup_base}
\end{table*}

\edit{
Next, we evaluate the applicability and utility of various self-supervised methods to serve as the basis for the discrete representation learning setup. 
Such analysis enables us to determine which self-supervised method can be utilized, allowing us to provide suggestions for specific scenarios.
For example, as discussed in \cite{haresamudram2022assessing}, simpler methods such as Autoencoders may be preferable--even though the performance is slightly lower--as they are easier and quicker to train.
Furthermore, Kmeans-based vector quantization (VQ) was also originally introduced in an Autoencoder setup \cite{van2017neural}.
Therefore, we not only study Autoencoders, but also other baselines such as Multi-task self-supervision and SimCLR, for their effectiveness towards functioning as the base for deriving discrete representations.
For this analysis we also perform brief hyperparameter tuning for the baseline methods, using the best parameters detailed in \cite{haresamudram2022assessing} and over the number of convolutional aggregator layers $\in \{2,4,6\}$ (similar to VQ-CPC).
The results from this analysis are given in Tab.\ \ref{tab:diff_self_sup_base}.
}

\edit{
First, we compare the performance of VQ-CPC against adding the VQ module to the Autoencoder setup (``VQ-Autoencoder'' in Tab.\ \ref{tab:diff_self_sup_base}).
For target datasets such as HHAR, Motionsense, and PAMAP2, the drop in performance while utilizing the convolutional Autoencoder as the base is around 10\%.
In the case of the remaining target scenarios, the reduction is lower at around 4\%, save for Myogym where the performance is comparable.
The established Multi-task self-sup.\ \cite{saeed2019multi} framework is ill-suited for discrete representation learning, with significant reductions in performance throughout, consistently performing worse by approx.\ 10\% for most datasets and peaking at around 29\% for Mobiact.
Similar analysis for SimCLR shows even more substantial reduction in performance, dropping by over 35\% consistently. 
As analyzed in Sec.\ \ref{sec:impact_of_encoder_arch}, the low performance of SimCLR and Multi-task self-sup.\ is likely due to the encoder architecture itself, which has a large receptive field (see below).
}

\edit{
In order to study whether smaller filters are more suitable for other self-supervised methods as well, we replace their encoders with the encoder network from VQ-CPC, and study the impact on the recognition accuracy.
For the Autoencoder, the effect is mixed, with the performance increasing slightly for datasets such as HHAR, Myogym, and PAMAP2, whilst reducing for Motionsense and MHEALTH. 
In the case of Multi-task self-sup., we observe that matching the encoder network (to VQ-CPC) has a significant impact on performance, resulting in improvements of approx.\ 8\% for PAMAP2, 13\% for HHAR, 24\% for Mobiact, and more modest 4-5\% for Motionsense and MHEALTH.
This clearly shows that large receptive fields such as the one resulting from the original Multi-task encoder are detrimental to discrete representation learning. 
Furthermore, utilizing a filter size of 4 is also a better option for other methods, including SimCLR.
Overall, we observe that the Autoencoder or Multi-task self-sup.\ with the replaced encoder can function as viable alternatives, albeit there is generally a reduction in performance relative to VQ-CPC. 
This can be useful in some situations as simpler methods may be preferable due to computational constraints.
}

\subsection{\edit{Potential Impact Beyond Standard Activity Recognition, and Next Steps}}
This work presents a \edit{\emph{proof-of-concept} in favor of learning discrete representations} for sensor-based human activity recognition.
Instead of applying state-of-the-art representation learning (via self-supervision) to learn dense, high-dimensional representations, we posit that learning discrete representations can result in comparable performance, while also potentially enabling more advanced tasks such as \edit{motif and activity discovery \cite{minnen2006discovering}}. 
\edit{
	We present an alternative data processing and feature extraction pipeline to the HAR community, to be utilized for further application scenarios but also to (once again) jumpstart research into developing discrete learning methods.
	In what follows, we describe future potential application scenarios where discrete representations can be especially useful. 
}

\paragraph{NLP-Based Pre-training:}
We observe in Tab.\ \ref{tab:adding_roberta} that adding pre-trained RoBERTa embeddings results in clear improvements over utilizing randomly initialized learnable embeddings for all target datasets.
For the locomotion-style and daily living datasets in particular, this results in state-of-the-art performance, which is highly encouraging, as it opens the possibility of adopting more powerful recent advancements from natural language processing for improved recognition of activities. 
\edit{
	Replacing RoBERTa, larger models such as GPT-2 \cite{radford2019language} or GPT-3 \cite{brown2020language} can be utilized on larger scale wearable sensor data such as UK Biobank \cite{doherty2017large}, leading to potential classification performance improvements.
	In addition, there is also potential to design and develop modifications to existing NLP-based pre-training, to make it more suitable for wearables applications, e.g., masking spans of tokens rather than randomly chosen individual ones for the masked language modeling \cite{joshi2020spanbert}, which has been shown to be more effective for time-series data \cite{baevski2019vq}.
	This is promising as advancements in NLP can also result in tandem improvements in sensor-based HAR.
	However, in situations where there are resource limitations, even inference with these models can be computationally prohibitive. 
	Therefore, recent works for miniaturizing and pruning the Transformer models \cite{lin2020pruning, sanh2019distilbert, lan2019albert} can be utilized to reduce the size while maintaining similar performance.
}

\paragraph{\edit{Activity Summarization:}}
\edit{
	Another avenue for leveraging our discrete representations involves utilizing established methods from NLP, which study unsupervised text summarization.
	They typically involve extracting key information/sentences from texts, i.e., are extractive rather than generative, as they do not have access to paired summaries for training.
	Approaches include utilized graphs \cite{mihalcea2004textrank, ramirez2021unsupervised} and clustering \cite{fung2003combining, gokhan2021extractive}.
	In recent years, large language models have been utilized for more accurate extractive summarization (e.g., \cite{ramirez2021unsupervised, gokhan2021extractive}).
	With our learned discrete representations, we can now utilize such summarization techniques in order to extract the most \emph{informative} sensor data, allowing us to, e.g., reduce noise (by removing unnecessary data), summarize the important movements during the hour/day etc.
	Therefore, summarization can help us understand routines better, by determining which portions of the day are most informative and therefore, representative of the activities through the day.
}

\paragraph{\edit{Sensor Data Compression:}}
\edit{
The discretization results in symbolic representations, which are essentially the `strings of human movements', effectively compressing the original data requiring substantially less memory for storing relative to multi-dimensional floating point numbers. 
This can be helpful in situations where data needs to be transmitted from the wearable to a mobile phone or server, leading to a reduction in transfer costs.
Furthermore, it also enables more efficient processing of extremely large-scale wearables datasets (such as the UK Biobank with 700k person days of data \cite{doherty2017large}), where the size is a crutch for analysis and model development.
}

\paragraph{\edit{Activity and Routine Discovery:}}
\edit{
	As mentioned in \cite{minnen2006discovering}, the process of discovering activities from unlabeled data is (in many ways) the opposite of building classifiers to recognize--known--activities using labeled data. 
	An important application includes health monitoring where typical healthy behavior can be characterized by such discovery algorithms, whereas they maybe difficult for humans (incl. experts) to fully specify \cite{minnen2006discovering}.
	One approach involves deriving `characteristic actions' via motif discovery, as such sequences are statistically unlikely to occur across activities and therefore correspond to important actions within the activity \cite{minnen2006discovering}. 
	Discovering motifs is easier in the discrete space (rather than raw sensor data space) as the simplification to a smaller alphabet aids with the identification of recurring patterns. 
	Especially for multi-channel data (such as accelerometry), our discretization method can be useful for discovering the characteristic actions across all three channels, without having to, for example, perform Principal Component Analysis to reduce the dimensionality to a single channel as in PERUSE \cite{oates2002peruse}.
	This can be highly useful for medical applications, e.g., recovery after injuries can be measured relative to pre-injury movements, or gait can be analyzed against typical expectations.
	Extending the idea to longitudinal studies, the routines can be discovered across days of data, providing insights into how movements and activities change over time. 
	Such setups can be vital for understanding and analyzing human behaviors.
}

\paragraph{Recognizing Fine-Grained Activities:}
In Tab. \ref{tab:diff_locs}, we see that utilizing an LSTM or GRU classifier for recognizing locomotion and daily living style activities at the wrist and waist is highly effective, resulting in the best (or similar to the best) performance for three benchmark datasets.
The improvement is obtained even though the discretization results in a loss of resolution due to the mapping to a small set of discrete symbols. 
However, this loss in resolution negatively impacts scenarios where the fine-grained activities need to be recognized (e.g., fine-grained gestures).
Therefore, \edit{a current limitation of the} proposed discretization \edit{method} is that it is not yet well-suited for applications, which require the discrimination between highly similar movements. 
\edit{
	Therefore, potential future work could include exploring Gumbel softmax vector quantization \cite{baevski2019vq, gumbel1954statistical} and additional losses that can promote diverse codebook usage via additional losses \cite{baevski2020wav2vec}.
	This could likely aid more fine-grained movements to be represented by their own symbolic representations, resulting in improved recognition.
}

\section{Conclusion}
The primary aim of this work was to serve as a \edit{\emph{proof-of-concept} to demonstrate how discrete representations can be learned from wearable sensor data, and that the performance of activity recognition systems based on such learned discretized representations is comparable to, if not better than, when using dense, i.e., continuous representations derived through state-of-the-art representation learning methods}.
In particular, we showed how automatically deriving the mapping between sensor data and a codebook of vectors in an unsupervised manner can solve some of the existing concerns with HAR applications based on discrete representations, including low activity recognition performance and difficulty with multi-channel data. 

A deeper dive into the workings of discretization showed that explicitly controlling the maximum dictionary size can result in better representations. 
Further, the addition of powerful NLP-based pre-training techniques such as RoBERTa resulted in improved activity recognition for all target datasets.
\edit{
Therefore, this paper casts the multi-channel time-series classification problem as a discrete sequence analysis problem (similar to natural language processing), thereby facilitating the adoption of recent advancements in discrete representation learning for the field of sensor-based human activity recognition.
}
In summary, our work offers \edit{an alternative feature extraction pipeline} in sensor-based HAR, allowing for discretized abstractions of human movements and therefore enables improved analysis of movements.

\bibliographystyle{ACM-Reference-Format}
\bibliography{refs}

\newpage
\section{Appendix}

\begin{figure}[!h]
	\centering
	\includegraphics[width=\textwidth]{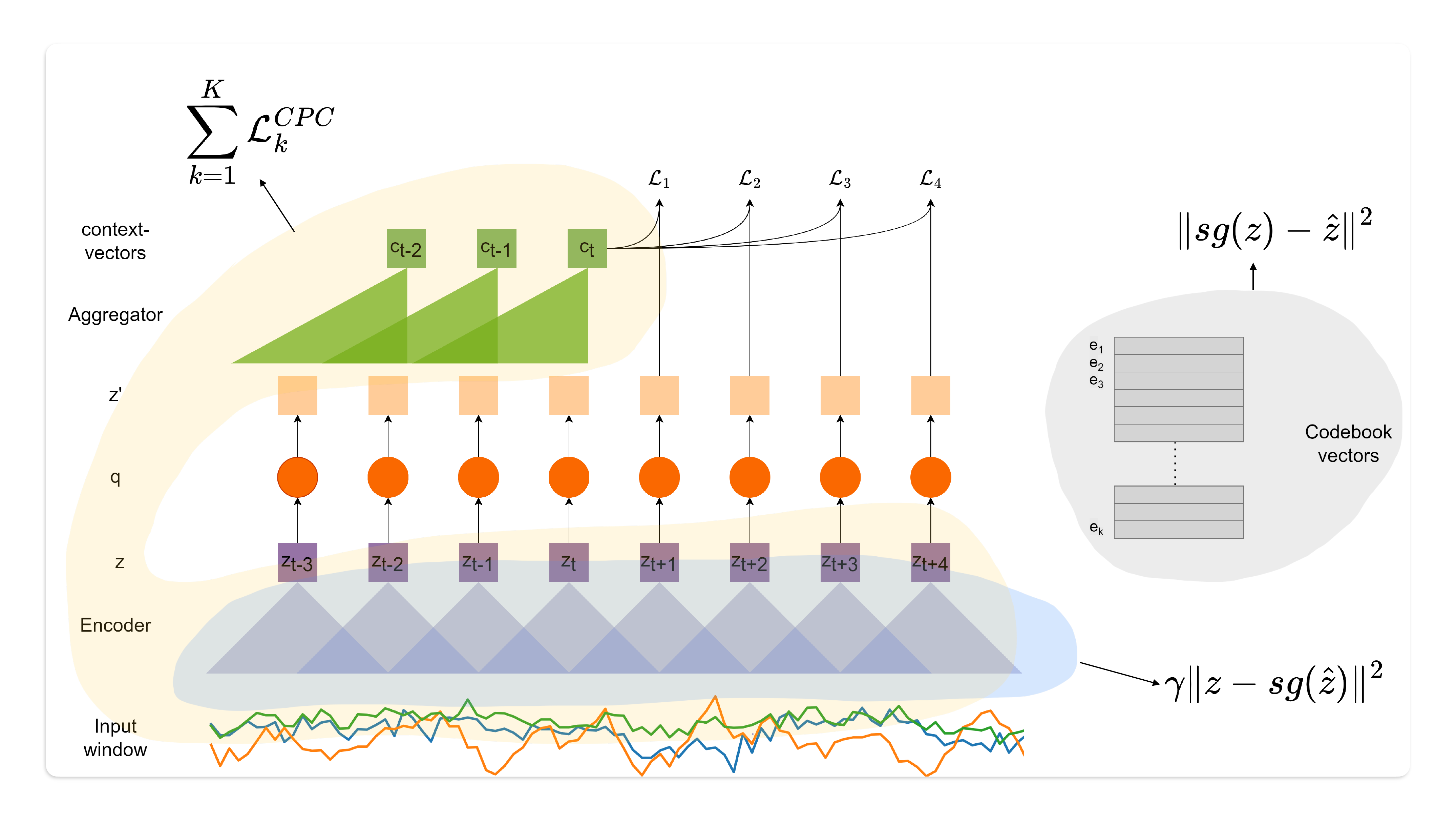}
	\caption{
		\edit{
		Visualizing how the weights of the VQ-CPC architecture are updated using different parts of the overall loss. 
		As detailed in Sec.\ \ref{sec:kmeans_quantization}, the Aggregator network is updated using $\sum_{k=1}^{K} \mathcal{L}^{CPC}_k$ whereas the codebook vectors utilize $ \Vert sg(z) - \hat{z} \Vert ^2 $.
		Finally, the encoder is updated using both  $\sum_{k=1}^{K} \mathcal{L}^{CPC}_k$ and $ \gamma \Vert z - sg(\hat{z}) \Vert ^2 $.
		}
	}
	\label{fig:vq_cpc_losses}
\end{figure}

\end{document}